\documentclass[pmlr,twocolumn,10pt]{jmlr} % W&CP article

\mlhtrack{findings}

\newif\iffinal
\finaltrue  % toggle this flag for camera-ready version after acceptance

\iffinal
    \ifmlhneedspmlr
      \jmlrvolume{XXX}
      \jmlryear{2026}
    \fi
    \ifmlhfindings \jmlrproceedings{}{ML4H 2026 - Findings Track}\fi
    \ifmlhdemo     \jmlrproceedings{}{ML4H 2026 - Demo Track}\fi
    \jmlrworkshop{Machine Learning for Health (ML4H) 2026}
\else
    \jmlrproceedings{}{Submitted to ML4H 2026: \mlhtrackname}
    \jmlrworkshop{Machine Learning for Health (ML4H) 2026}
\fi

\usepackage{rotating}% for sideways figures and tables
\usepackage{longtable}% for long tables

\usepackage{booktabs}
\usepackage{siunitx}

\usepackage[switch]{lineno}

\title{Predicting days alive and out of hospital using ordinal-binning gradient-boosting regression}

\author{
\Name{Mathias W. Roesler} \Email{mathias.roesler@auckland.ac.nz}\\
\addr Department of Surgery, University of Auckland, Auckland, New Zealand
\AND
\Name{Luke Boyle} \Email{luke.boyle@TeWhatuOra.govt.nz} \\
\addr Department of Anaesthesiology, University of Auckland, Auckland, New Zealand
\AND
\Name{Malek Connor} \Email{malek.connor@gmail.com} \\
\addr University of Canterbury, Christchurch, New Zealand  
\AND
\Name{Leo Y. Li-Han} \Email{li-han.leo@mayo.edu} \\
\addr Department of Surgery, Mayo Clinic, Rochester, Minnesota, USA 
\AND
\Name{Cameron Wells} \Email{cameron.wells@auckland.ac.nz} \\
\addr Wellington Regional Hospital, Wellington, New Zealand
\AND
\Name{Chris Varghese} \Email{chris.varghese@auckland.ac.nz} \\
\addr Department of Surgery, University of Auckland, Auckland, New Zealand
}

\begin{document}

\maketitle

\ifmlhdemo\else
\begin{abstract}
Days alive and out of the hospital (DAOH) is a patient-centred quality outcome that reflects recovery and healthcare utilisation after surgery. It is a highly non-linear, discrete prediction target with multifactorial contributors, and traditional regression models cannot capture the zero-inflation in the DAOH distribution. 

We designed a novel ordinal-binning gradient-boosting regression model (OrdBoost) to predict DAOH at 90 days. The model generated prediction intervals associating an uncertainty measure for each prediction. OrdBoost was compared against other regression models and an NGBoost model, also capable of natively producing prediction intervals. All models were trained on over 2.2 million adults undergoing surgery in Aotearoa, New Zealand.

Overall, the OrdBoost model offered excellent performance, with a mean absolute error (MAE) of 3.86 days (95\% CI [3.81, 3.91]) and a continuous ranked probability score (CRPS) of 2.95 days (95\% CI [2.91, 2.98]). The prediction intervals were sharp, with a 26-day average interval width for the 95\% PI with empirical coverage of 97\% for a nominal coverage of 95\%. Moreover, OrdBoost captured the data spike at 0 days. 

The OrdBoost model accurately predicted DAOH with prediction intervals to measure uncertainty, demonstrating solid performance for prediction of targets with complex, non-linear data distributions.
\end{abstract}
\fi

\ifmlhneedsstatements
\paragraph*{Data and Code Availability}
The data used to train and test the model were from the New Zealand National Minimum Dataset (NMDS), which can be obtained from the New Zealand Ministry of Health National Collections team. Requests can be made by contacting data-enquiries@health.govt.nz.

Reproducibility and fitting were performed with the \texttt{medpipe} Python package freely available on GitHub \href{https://github.com/Surgical-Recovery-and-Safety-Lab/medpipe}{here}. 
A Python package for the OrdBoost model, named \texttt{ordboost} was created and is available on PyPI. The source code is freely available on GitHub \href{https://github.com/Surgical-Recovery-and-Safety-Lab/ordboost}{here}. 
All the code written to produce figures and tables is freely available on GitHub  \href{https://github.com/Surgical-Recovery-and-Safety-Lab/daoh-prediction}{here}. 

\paragraph*{Institutional Review Board (IRB)}
Ethical approval was obtained from the Auckland Health Research Ethics Committee (AH29056).
\fi

\section{Introduction}
\label{sec:intro}
Days at home, or days alive and out of hospital (DAOH), is a patient-centred outcome that measures the number of days a patient spends alive and out of hospital after surgery over a given temporal horizon \citep{brown_patient-centred_2025}. DAOH is a composite outcome comprising mortality, hospital length of stay, readmission to hospital, readmission length of stay, and discharge to any care or rehabilitation facility providing hospital-level care \citep{brown_patient-centred_2025}. Thus, DAOH captures the risk of readmission or prolonged stay due to post-operative complications, and can be used as an indicator of post-operative recovery. Some studies assigned a DAOH value of 0 for any patient who died within the observation period \citep{myles_validation_2017,jerath_days_2019}, regardless of the number of days actually spent out of the hospital. Other studies \citep{wells_hospital_2026, brown_evaluation_2026}, however, do not make that distinction and define DAOH at a temporal horizon $T$, as adopted in this study, as:

\begin{equation}
\label{eq:daoh}
DAOH_T =  \begin{cases}
    D_{T} - LOS_{0} - \sum_i^NLOS_i \\
    0, \text{if } LOS_0 - \sum_i^NLOS_i > D_T
\end{cases}
\end{equation}
where $LOS_{0}$ is the initial length of stay after surgery, which is at least 1, $LOS_i$ is the length of stay of the i\textsuperscript{th} readmission, and $D_{T}$ is the number of days alive at horizon $T$. Thus, $DAOH_{T}$ decreases every time the patient is readmitted to the hospital. From \equationref{eq:daoh}, there are two mechanisms to obtain a $DAOH_{T}$ value of 0: the patient died in hospital, in that case $D_T = LOS_0$ and $N = 0$, or they spent more than $T$ days in hospital without being discharged. If a patient does not die, then $D_T = T$. The temporal horizon can be set to any value, with common choices being one month (30 days), three months (90 days), six months (180 days), or one year (365 days) \citep{boyle_using_2025,spurling_validation_2022, brown_evaluation_2026}. However, an evaluation of DAOH across different temporal horizons showed that 90 days was a better choice than 30 days due to floor effects, which reduced the discriminative power of the outcome for survivors with longer hospital admissions among intensive care patients \citep{brown_evaluation_2026}. 

DAOH has been used recently in retrospective studies to evaluate healthcare quality assessment \citep{boyle_using_2023, wells_hospital_2026}, equity of outcomes after surgery \citep{boyle_using_2025}, and the impact of socioeconomic status \citep{austin_socioeconomic_2020}. The use of this metric as a clinical outcome has been validated across diverse settings, including older patients undergoing surgery for hip fractures \citep{wu_evaluation_2022} and those undergoing emergency laparotomy \citep{spurling_validation_2022}. Hence, this validated patient-centred clinical outcome metric is valuable for guiding patient expectations preoperatively and represents a more nuanced measure of recovery than crude metrics such as mortality \citep{brown_patient-centred_2025}. 

Models predicting DAOH have been developed using various strategies, including a multistate model \citep{bottle_regional_2018}, machine learning models \citep{ribeiro_predicting_2026}, and statistical models \citep{grudzinski_comparing_2022}. However, the zero-inflated and left-skewed distribution of DAOH makes accurate predictions more challenging \citep{holodinsky_comparing_2021}. \citet{holodinsky_comparing_2021} compared multiple regression models and found that machine learning models outperformed statistical models; however, no model effectively captured the zero-inflated and skewed nature of the DAOH distribution. 

We present a novel method for predicting DAOH following surgery under general or neuraxial anesthesia using an ordinal-binning gradient-boosting strategy. The target values were first encoded into ordinal bins, and an ensemble of binary classifiers was trained for each cumulative edge threshold. Subsequently, a bin mapper transforms the predicted probabilities into cumulative distribution functions. This novel approach extracts prediction intervals for each patient, aiding model output interpretability for clinical care planning.

\section{Material and Methods}
\label{sec:methods}
\subsection{Data}
This was a national, population-based retrospective sample of 2,585,586 adults who underwent surgery under general or neuraxial anesthesia in public hospitals in Aotearoa, New Zealand, between 1 January 2010 and 31 December 2024. 

A total of 11 features were extracted from the data (presented in \tableref{tab:patient_chars}). The data contained six categorical features (sex at birth, ethnicity, admission source, admission acuity, specialty, and sub-specialty), one numerical feature (age), two ordinal features (American Society of Anesthesiologists (ASA) physical status classification and operation severity), and two binary features (prior cancer and trauma). The data were split into three distinct sets based on the year of operation: \begin{enumerate}
    \item Train set (operation years: 2010 -- 2022), constituted the training data to fit the model and contained 2,239,441 patients (representing 86.6\% of the total data);
    \item Validation set (operation year: 2023), was used to fine-tune the components of the OrdBoost model (see \appendixref{sec:n_bins}, \appendixref{sec:bin_dist}, and \appendixref{sec:mappers}) and contained 177,395 patients (representing 6.9\% of the total data) 
    \item Test set (operation year: 2024) was held out for testing the model and contained 168,750 patients (representing 6.5\% of the total data).
\end{enumerate}

\subsection{Preprocessing}
Minimal preprocessing was performed before training the model. The categorical features were encoded to integers using the \texttt{sklearn OrdinalEncoder}. In total, 541,520 values were missing across all three sets. These were mainly due to missing ASA scores (486,219, 34,711, and 17,679 for the training, validation, and test sets, respectively), with only 2,850 (2,377, 221, and 252 for the training, validation, and test sets, respectively) missing ethnicity values, and 61 (16, 20, and 25 for the training, validation, and test sets, respectively) missing sex values. Rather than imputing the missing values, a model capable of handling missing data were used (see \sectionref{sec:model} for more details). 

\subsubsection{Prediction target}
The prediction target was days alive and out of hospital for $T=90$ days ($DAOH_{90}$). This outcome represented the number of days spent alive and out of the hospital three months after surgery. By definition, the day of surgery is always considered a day in the hospital; therefore, $DAOH_{90}$ values range from 0 to 89 days. From \equationref{eq:daoh}, a $DAOH_{90}$ of 0 corresponds to a patient who died in hospital or had a post-operative length of stay of at least 90 days, and a $DAOH_{90}$ of 89 to a patient who was discharged immediately after surgery. 

Because $DAOH_{90}$ exhibited a complex distribution with a point mass at $y=0$ and a hard limit at $y=89$, the target $DAOH_{90} = Y \in \{0, 1, \dots, 89\}$ was discretized into $K = 12$ ordinal bins, $\mathcal{B} = \{b_0, b_1, \dots, b_{K-1}\}$. The bin boundaries were chosen based on empirical data density to capture zero-inflation, one and two-month stays in hospital, and high-density upper ranges:
\begin{align*}
b_0 &= \{0\}, & &b_1 = [1, 30], & &b_2 = [31, 60], \\
b_3 &= [61, 70], & &b_4 = [71, 75], & &b_5 = [76, 79], \\
b_6 &= [80, 84], & &b_7 = \{85\}, & &b_8 = \{86\}, \\
b_9 &= \{87\}, & &b_{10} = \{88\}, & &b_{11} = \{89\}. \\
\end{align*}

\subsection{OrdBoost model}
The ordinal-binning gradient-boosting regression (OrdBoost) model comprised two components: 
\begin{enumerate}
    \item the ordinal classifier which fitted an ensemble of 11 binary classifiers for the $K = 12$ bins, and
    \item the bin mapper that converted the probabilities into continuous probability distributions.
\end{enumerate}

Ablation studies were performed to evaluate the effects of each component in the OrdBoost model. Three studies were conducted to evaluate the effect of the number of bins (see \appendixref{sec:n_bins}), the distributions of the bins (see \appendixref{sec:bin_dist}), and finally the mapping strategy (see \appendixref{sec:mappers}).

\subsubsection{Ordinal classifier}
\label{sec:model}
The ordinal classifier modelled the probability distribution over ordered bins $b_0 < b_1 < \dots < b_{K-1}$. Unlike standard multiclass classification, the ordinal target labels preserved proximity: outcome $b_0$ is closer to $b_1$ than to $b_{11}$.

To incorporate this ordinal structure, an ensemble of $K - 1$ binary base classifiers was trained using a cumulative binary edge formulation. Let $c_k = \max(b_k)$ denote the upper bound threshold of bin $b_k$ for $k \in \{0, 1, \dots, K-2\}$. For a feature vector $\mathbf{x}$, the $k$-th binary classifier estimates the cumulative probability $F_k(\mathbf{x}) = P(Y \le c_k \mid \mathbf{x})$, separating instances where $Y \in \bigcup_{j=0}^k b_j$ from $Y \in \bigcup_{j=k+1}^{K-1} b_j$. Each binary edge model was implemented using histogram-based gradient-boosted decision trees (\texttt{HistGradientBoostingClassifier}).

Because the $K-1$ binary classifiers were trained independently, the raw
predicted edge probabilities $\hat{F}_k(\mathbf{x})$ may violate the
monotonicity required of a cumulative distribution function (CDF),
$\hat{F}_k(\mathbf{x}) \le \hat{F}_{k+1}(\mathbf{x})$. Let
$\boldsymbol{\hat{F}}(\mathbf{x}) = [\hat{F}_0(\mathbf{x}), \dots,
\hat{F}_{K-2}(\mathbf{x})]^T$ denote the raw edge probabilities of a patient.
To enforce monotonicity, $\hat{F}_k(\mathbf{x})$ was regressed on the edge
index $k$ under a non-decreasing constraint independently for each patient. This was the least-squares
projection of $\boldsymbol{\hat{F}}(\mathbf{x})$ onto the set of
non-decreasing sequences, computed with the pool-adjacent-violators
algorithm. The result was
$\boldsymbol{\tilde{F}}(\mathbf{x}) = [\tilde{F}_0(\mathbf{x}), \dots,
\tilde{F}_{K-2}(\mathbf{x})]^T$ with $\tilde{F}_k(\mathbf{x}) \le
\tilde{F}_{k+1}(\mathbf{x})$ for all $k$. Sequences that were already non-decreasing were returned unchanged. The boundary values were then set to
$\tilde{F}_{-1}(\mathbf{x}) = 0$ and $\tilde{F}_{K-1}(\mathbf{x}) = 1$.

The discrete probability mass function (PMF) across bins, $p_k(\mathbf{x}) = P(Y \in b_k \mid \mathbf{x})$, was subsequently obtained via forward differences:
\begin{equation}
p_k(\mathbf{x}) = \tilde{F}_k(\mathbf{x}) - \tilde{F}_{k-1}(\mathbf{x}), \text{ for } k = 0,\dots, K-1.
\end{equation}

The \texttt{HistGradientBoostingClassifiers} were all trained with the following heuristically identified parameters: loss: \texttt{log loss}, learning rate: 0.1, maximum number of iterations: 100, maximum number of leaves: 512, minimum number of samples per leaf: 512. All other parameters were left at their default values. Early stopping was implemented using 10\% of the training data, randomly selected, as a validation set. Training MAE and CRPS (see \sectionref{sec:acc}) were evaluated with 10-fold cross-validation, and results were presented in \appendixref{sec:train}.

\subsubsection{Bin mapper}
\label{sec:mapper}
The bin mapper transformed the discrete PMF $p_k(\mathbf{x})$, for $k \in \{0, 1, \dots, K-1\}$, into a continuous predictive CDF $F(y \mid \mathbf{x})$. The mapper created a grid of points and associated weights at which to evaluate the CDF. The grid of points at which the CDF was evaluated was created based on the bin edges $b_0, \dots, b_{11}$, and used a quantile strategy. Let $Q_k(q)$ denote the empirical $q$-quantile of the training outcomes within bin $b_k$, for $q\in \{0, 0.10, 0.25, 0.50, 0.75, 0.90, 1\}$. For patient $x$, let $S_k(x) = \Sigma_{j<k} p_j(x)$ be the cumulative probability before bin $k$. The CDF value at each evaluation point $q$ within a bin, namely outcome knots, is given by
\begin{equation}
    F(Q_k(q) | x) = S_k(x) + q\cdot p_k(x)
\end{equation}

The predictive within-bin CDF was obtained by linearly interpolating the cumulative probability between adjacent unique outcome knots. Therefore, patient-wise characteristics determined the mass assigned to each bin, while the within-bin mass distribution was determined by the empirical training distribution shared across patients. When quantiles were identical, duplicate grid points were removed, and the largest $q$ was retained. For single-width bins, $q$ was always set to 1.

The first grid point was set to $-1\times10^{-4}$ with a weight of 0 to ensure the first element of the CDF evaluates to 0. The second point was the single-value bin $b_0 = 0$ with weight 1. Finally, an additional grid point was inserted at $89 - 1\times10^{-4}$ with weight $K - 1 + \bar{q}$, where $\bar{q}$ denotes the empirical fraction of training targets within $b_{K-1}$ falling strictly below 89. This point, together with the existing terminal boundary point at $y = 89$ (weight $K$, mapped to $F = 1$), represents the substantial point mass of patients discharged with no further hospital utilisation ($DAOH_{90} = 89$) as a steep, near-discontinuous rise across a negligible interval, rather than allowing this probability mass to be smoothed uniformly across the full width of $b_{K-1}$ by the standard intra-bin quantile treatment described above. 

\subsection{Comparison models}
The OrdBoost model was compared against other models. The first was a naive baseline to contextualise some of the evaluation metrics against a reference forecast. Two simple baseline machine learning models (a histogram gradient-boosting regressor and an ordinal ridge regressor) were also used; however, they were unable to produce CDFs. Thus, a final model, a natural gradient-boosting model, was the main point of comparison since it could produce CDFs.

\subsubsection{Naive baseline}
The naive population baseline model (referred to as the naive model) was constructed from the unconditional empirical distribution of $DAOH_{90}$ in the training set, independent of patient features:
\begin{equation}
    F_{\text{base}}(y) = \frac{1}{N_{\text{train}}}\sum_{i=1}^{N_{\text{train}}}
    \mathbf{1}(y_i \le y),
\end{equation}
where the same $F_{\text{base}}$ was applied identically to every test patient, representing the best achievable forecast in the absence of any covariate information. 

\subsubsection{Natural gradient-boosting}
The ordinal-binning gradient-boosting model was compared to another regression model capable of producing CDFs: a natural gradient-boosting regressor (NGBoost) \citep{duan_ngboost_2020}. The NGBoost model was fitted to the non-binned $DAOH_{90}$ data with a learning rate of 0.1; other parameters were left at their default values. Similarly to the OrdBoost model, early stopping was implemented if the validation loss did not improve for 10 iterations. The training MAE and CRPS (see \sectionref{sec:acc}) were evaluated using 10-fold cross-validation, and the results are presented in \appendixref{sec:train}.

The NGBoost model had no constraints on its predictions; therefore, predictions were clipped to the [0--89] range of the target.

\subsubsection{Histogram gradient-boosting regressor}
The histogram gradient-boosting regressor (HGBRegressor) was implemented with the \texttt{sklearn} package with an absolute error loss function and learning parameters similar to those of the OrdBoost classifiers (learning rate: 0.1, maximum number of iterations: 100, maximum number of leaves: 512, minimum number of samples per leaf: 512). This model represented the performance achieved by a simple regression model. Similar to the NGBoost model, the predictions from the HGBRegressor model were clipped. 

\subsubsection{Ordinal ridge regressor}
A simple ordinal ridge regression model (OrdinalRidge) included the ordinal information from the data. Additional data preprocessing was performed for the OrdinalRidge model because it does not natively handle missing values. The missing ASA, sex, and ethnicity values were imputed using \texttt{sklearn}'s \texttt{SimpleImputer}. The sex, trauma, prior cancer, admission source, admission acuity, and ethnicity features were one-hot encoded. The surgical specialties and sub-specialties had too many possible values to be one-hot encoded; therefore, a target encoder was used. Finally, age, ASA scores, operation severity, and both encoded targets were standardised with \texttt{sklearn}'s \texttt{StandardScaler}. The predictions were also clipped to the correct $DAOH_{90}$ range.

\subsection{Evaluation}
\label{sec:eval}
Model performances were evaluated on the full held-out test set (N = 168,750) for accuracy, continuous calibration, and prediction interval quality. The 95\% confidence intervals were computed for each metric using bootstrapping with 1,000 iterations.

\subsubsection{Accuracy, scoring rules, and skill scores}
\label{sec:acc}
Point-prediction accuracy was evaluated by computing the Mean Absolute Error (MAE) between the predicted median value $F^{-1}(0.5)$ and the ground-truth outcome for all models.
The Continuous Ranked Probability Score (CRPS) is a proper scoring rule for evaluating CDFs \citep{gneiting_probabilistic_2007}. Let $F$ be the CDF such that $F(y) = P(Y \le y)$, then the CRPS is defined as: 
\begin{equation}
    CRPS(F, y_{n}) = \int_{-\infty}^{\infty}\left(F(y) - \mathbf{1}(y \ge y_{n})\right)^2dy,
\end{equation}
where $y_n$ is the true observation, and $\mathbf{1}$ is the Heaviside function, defined as:
\begin{equation}
\mathbf{1}(y \ge y_{n})=
    \begin{cases}
        1 & \text{if } y \ge y_{n} \\
        0 & \text{otherwise }
    \end{cases}
\end{equation}

Both the MAE and the CRPS provide global estimates for the accuracy. The CRPS skill score was computed to evaluate the model against the naive model, and was defined as
\begin{equation}
    CRPSS = 1 - \frac{\overline{CRPS}_{model}}{\overline{CRPS}_{base}}, 
    \\
\end{equation}
where $\overline{CRPS}_{base}$ denoted the mean CRPS obtained by scoring $F_{base}$ against the true outcomes, and $\overline{CRPS}_{model}$ represented the mean CRPS of the model. Similarly, the MAE skill score (MAESS) was defined as
\begin{equation}
    MAESS = 1 - \frac{MAE_{model}}{MAE_{base}}
\end{equation}
where $MAE_{base}$ was the MAE if the model only predicted the median $DAOH_{90}$ of the training set.
A skill score of 0 indicates performance equal to the naive model, 1 indicates a perfect forecast, and negative values indicate performance worse than the baseline. 

\subsubsection{Continuous calibration diagnostics}
\label{sec:pit}
Statistical calibration was evaluated for the OrdBoost and NGBoost models via the Probability Integral Transform (PIT) histogram, following the methods from \citet{taggart_assessing_2022} to account for the discontinuities at 0 and 89. A calibrated model produces a uniform distribution. To measure the deviation from the uniform distribution, the $\alpha$ score was computed \citep{renard_understanding_2010}.

Marginal calibration \citep{gneiting_probabilistic_2007} was assessed by plotting the difference between the empirical cumulative distribution function:
\begin{equation}
    \text{eCDF}(y) = \frac{1}{N}\sum_{n=1}^{N}\mathbf{1}(y_{n} \le y),
\end{equation}
and the mean predictive CDF across all test samples:
\begin{equation}
    \overline{\text{CDF}}(y) = \frac{1}{N}\sum_{n=1}^{N}{F}_{n}(y \mid \mathbf{x}_{n}),
\end{equation}
for $y \in \{0, 1, \dots, 89\}$, where $\mathbf{x}_{n}$ denotes the feature vector, $y_{n}$ the true outcome, and $F_{n}(y \mid \mathbf{x}_{n})$ the predicted CDF for test sample $n$.

\subsubsection{Prediction interval evaluation}
Prediction intervals from the OrdBoost and NGBoost models were evaluated to assess empirical coverage and sharpness across nominal confidence levels $(1 - c)$, where $c \in \{0.05, 0.10, \dots, 0.90\}$. Coverage measures the proportion of observations falling within the predicted interval, whereas sharpness measures interval width concentration. A performant model must exhibit both nominal coverage and high sharpness \citep{gneiting_probabilistic_2007}. 

\section{Results}
\label{sec:res}
\begin{figure}[h]
\floatconts
  {fig:distributions}
  {\caption{Days alive and out of hospital capped at 90 days ($DAOH_{90}$) distributions. The top panel shows the original $DAOH_{90}$ distribution for the test set. The bottom panel displays the $DAOH_{90}$ distribution after the binning process used by the OrdBoost model.}}
  {\includegraphics[width=\linewidth]{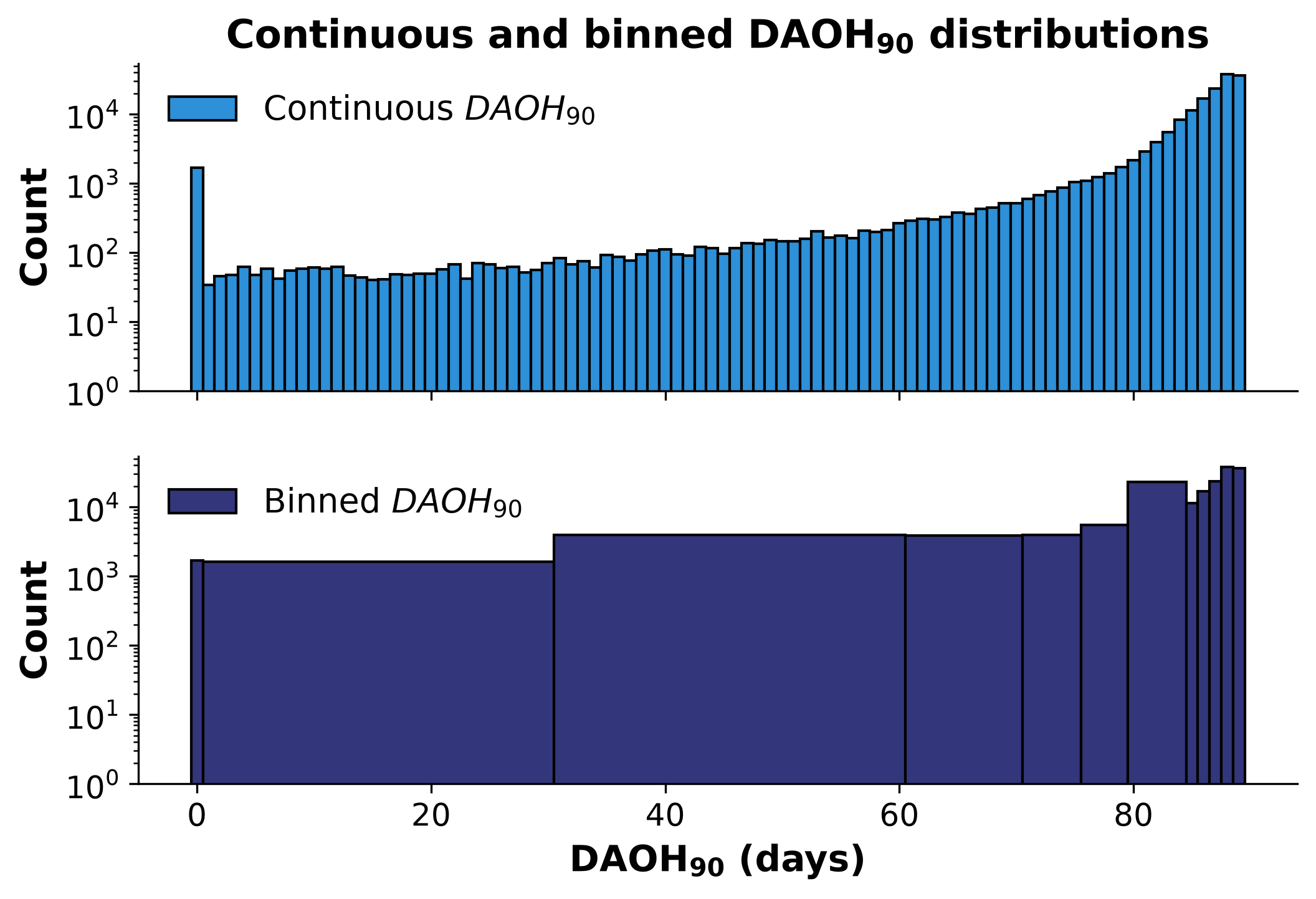}
  }
\end{figure}

\figureref{fig:distributions} shows the $DAOH_{90}$ distributions before and after the binning process on a log-scale. The top panel clearly showed the spike at 0 and the large number of patients who spent less than a week in hospital after surgery. In total, 36,010 (21.3\% of the test set) of patients had a $DAOH_{90}$ of 89 and 1,698 patients had a $DAOH_{90}$ of 0, representing 1.0\% of the test set. This number contrasted against the 34 patients with a $DAOH_{90}$ value of 1 and highlighted the zero-inflation of the $DAOH_{90}$ distribution. 

\tableref{tab:tab1} presented the MAE, MAESS, CRPS, and CRPSS for all four models on the test set. The two simple regression models (HGBRegressor and OrdinalRidge) did not produce CDFs; therefore, the CRPS was reduced to the MAE. The OrdinalRidge model was the least performant, with negative MAESS and CRPSS, indicating worse performance than the naive model. In contrast, the HGBRegressor outperformed the NGBoost model, with a lower MAE of 3.92 days (95\% CI [3.87, 3.97]) and nearly identical CRPS and CRPSS. The OrdBoost model had a similar MAE to the HGBRegressor model (3.86, 95\% CI [3.81, 3.91] days) but a better CRPS of 2.95 (95\% CI [2.91, 2.98] days) and a CRPSS approximately 4.2 times larger.

\begin{sidewaystable}[htbp]
\floatconts
  {tab:tab1}
  {\caption{Accuracy and scoring rules comparison between all models with 95\% CI.}}
  {\begin{tabular}{lcccc}
\toprule
\textbf{Metrics} & \textbf{HGBRegressor} & \textbf{OrdinalRidge} & \textbf{NGBoost} & \textbf{OrdBoost} \\
\midrule
\textbf{MAE (days)} & 3.92 [3.87, 3.97] & 6.19 [6.15, 6.24] & 5.06 [5.01, 5.10] & 3.86 [3.81, 3.91] \\
\textbf{MAESS} & 0.221 [0.217, 0.225] & -0.233 [-0.241, -0.224] & -0.007 [-0.014, 0.000] & 0.232 [0.228, 0.237] \\
\textbf{CRPS (days)} & 3.92 [3.87, 3.97] & 6.19 [6.15, 6.24] & 3.86 [3.83, 3.90] & 2.95 [2.91, 2.98] \\
\textbf{CRPSS} & 0.072 [0.067, 0.076] & -0.469 [-0.478, -0.458] & 0.084 [0.078, 0.091] & 0.301 [0.297, 0.306] \\
\bottomrule
\end{tabular}
}
\end{sidewaystable}

\figureref{fig:pit} presented the PIT histograms for the NGBoost and OrdBoost models. The NGBoost PIT histogram was hump-shaped, with a concentration of values near the centre, signalling model miscalibration. The hump shape is indicative of an overdispersed model, as evidenced by overly wide prediction intervals and an underconfident model. The OrdBoost PIT histogram also deviated from the uniform distribution; however, the nature of the miscalibration differed. The PIT values accumulated towards the right side of the histogram as a \begin{figure}[h]
\floatconts
  {fig:pit}
  {\caption{PIT histogram for both NGBoost and OrdBoost models. The NGBoost model was over-dispersed with a hump-shaped PIT histogram. In contrast, the OrdBoost model had a biased, asymmetric PIT histogram consistent with under-prediction. Error bars and bracketed numbers represented 95\% CI.}}
  {\includegraphics[width=\linewidth]{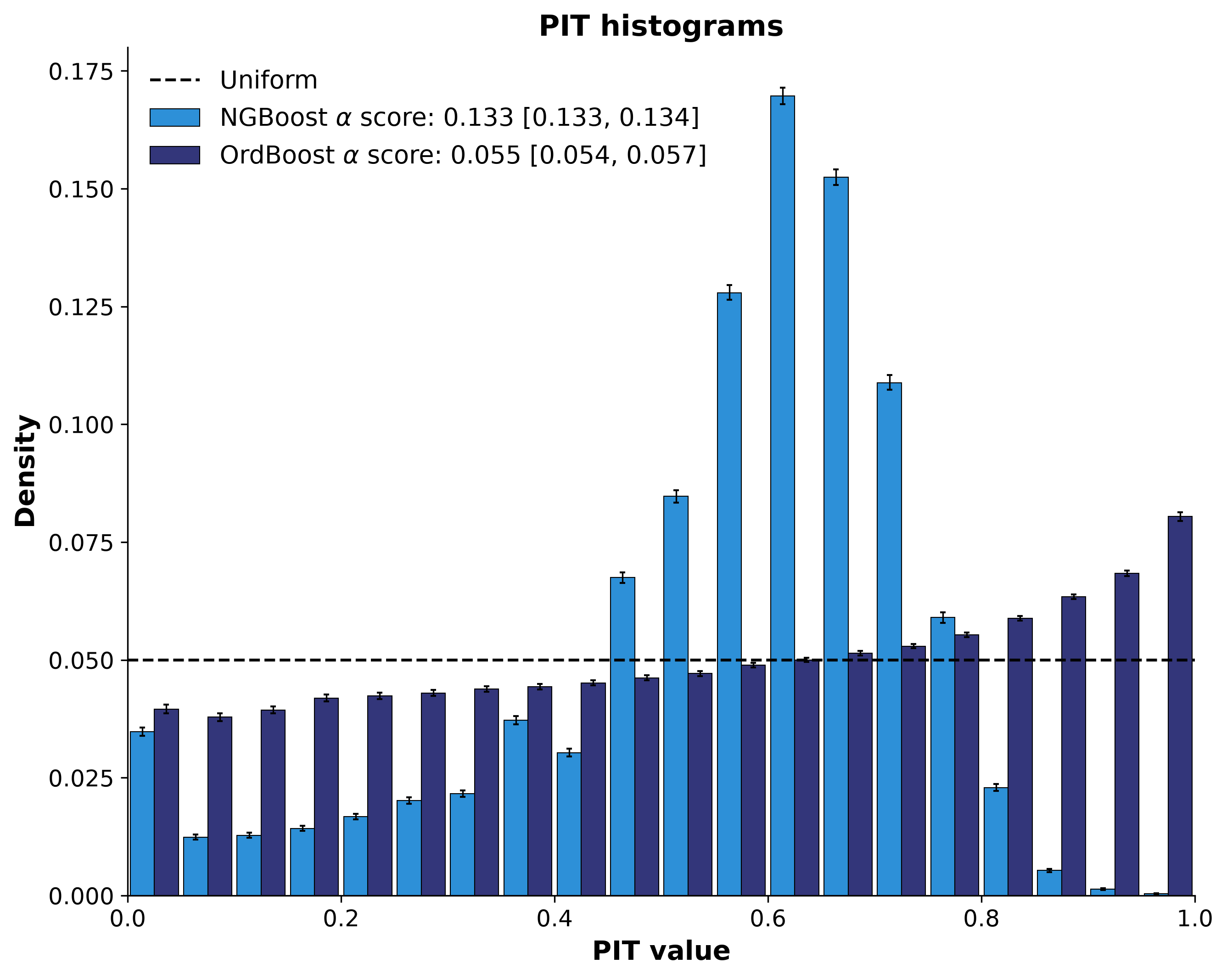}
  }
\end{figure}consequence of the model under-predicting the $DAOH_{90}$ values. Overall, the OrdBoost model had a PIT 
histogram closer to the uniform distribution; consequently, the $\alpha$ score was lower at 0.055 (95\% CI [0.054, 0.057]) compared to 0.133 (95\% CI [0.133, 0.134]) for the NGBoost model. Thus, the miscalibration was greater for the NGBoost model than for the OrdBoost model.

\figureref{fig:cal} shows the marginal calibration of the models. These results were in agreement with those from \figureref{fig:pit}. Indeed, the OrdBoost model showed a negative difference between the empirical CDF and the mean CDF, confirming that it tended to predict lower $DAOH_{90}$ values than expected. The NGBoost marginal calibration varied more, with a positive curve for $DAOH_{90} \le 40$ and $DAOH_{90} \ge 85$, but a negative curve otherwise. Thus, on average, the NGBoost model predicted larger values for patients with a true $DAOH_{90}$ value less than 40 or greater than 85, but predicted lower values in between.

\begin{figure}[hbp]
\floatconts
  {fig:interval}
  {\caption{Prediction interval evaluation for both models.}}
  {%
    \subfigure[For both models, the prediction intervals over-covered with empirical coverage greater than the nominal coverage.]{\label{fig:coverage}%
      \includegraphics[width=\linewidth]{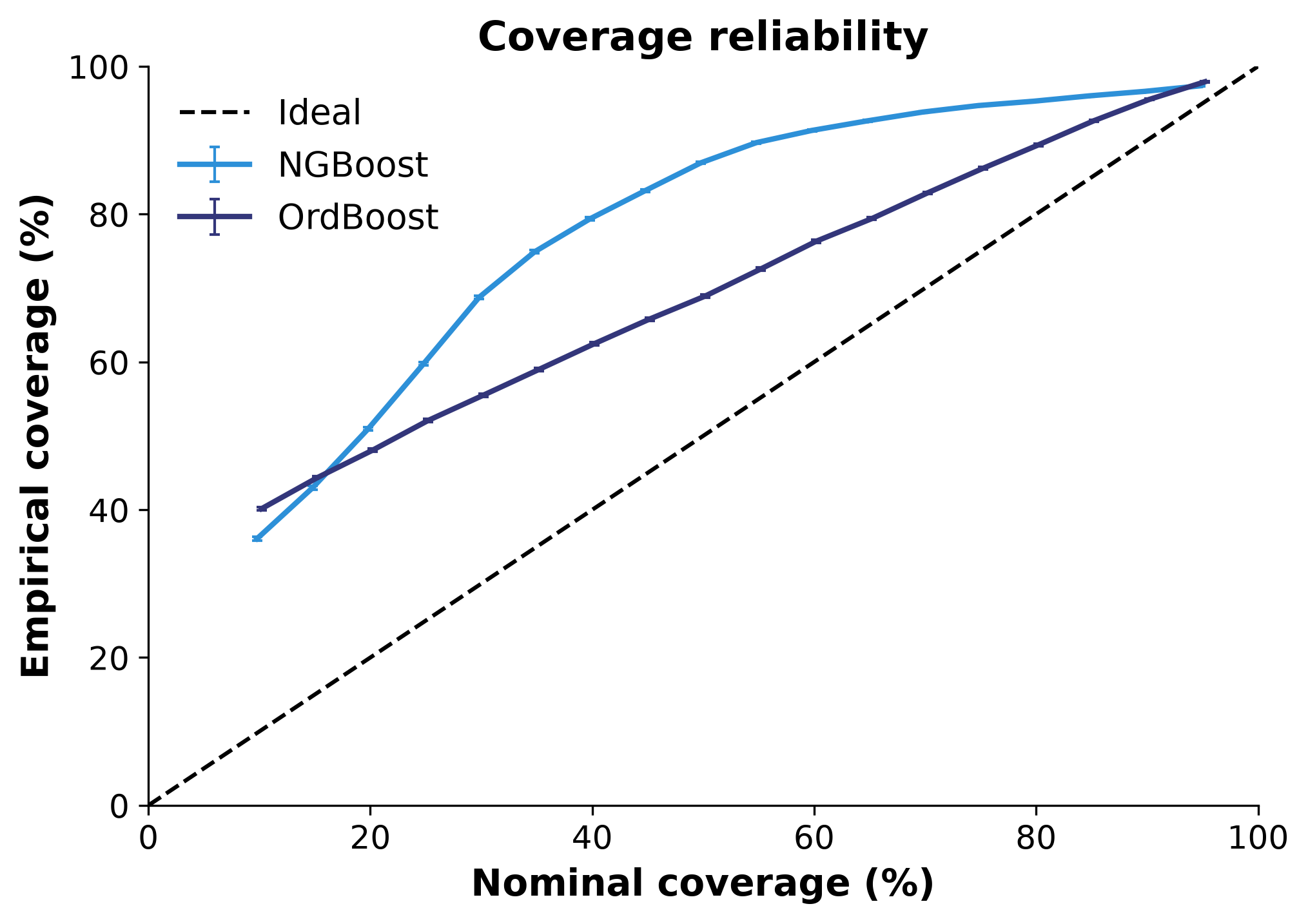}}%
    \qquad
    \subfigure[The mean interval width increased with the coverage level. The OrdBoost model produced tighter intervals than the NGBoost model, except at 95\% nominal coverage. Error bars represented 95\% CI.]{\label{fig:sharpness}%
      \includegraphics[width=\linewidth]{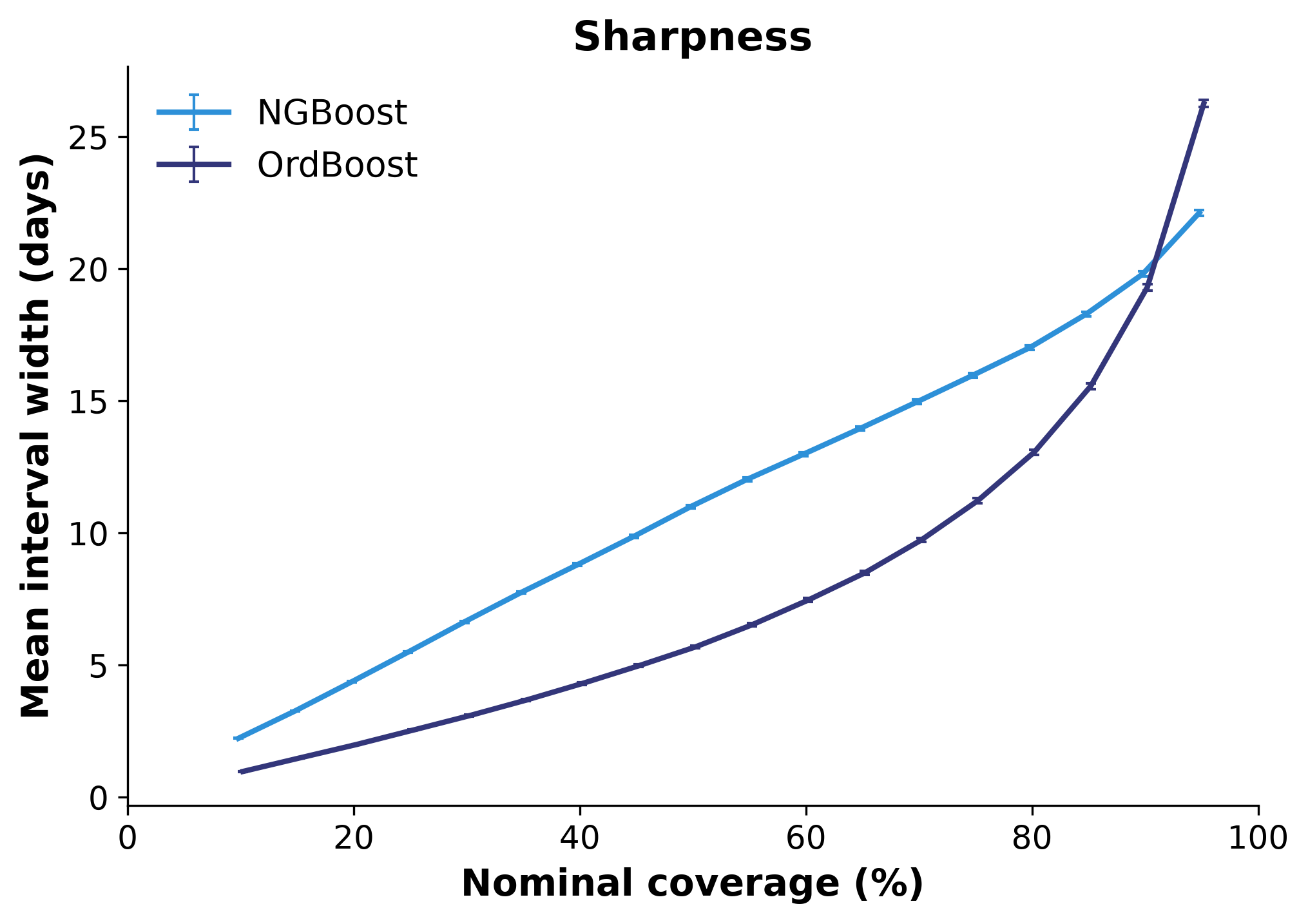}}
  }
\end{figure}

\figureref{fig:interval} shows the evaluation of the prediction intervals for both models at different nominal coverage levels. The coverage reliability (\figureref{fig:coverage}) 
for both models deviated from ideal coverage. The models exhibited higher-than-expected coverage \begin{figure}[h]
\floatconts
  {fig:cal}
  {\caption{Marginal calibration curves for the NGBoost and OrdBoost models. The OrdBoost model showed smaller miscalibration over the $DAOH_{90}$ range, while the NGBoost model showed miscalibration in both directions.}}
  {\includegraphics[width=\linewidth]{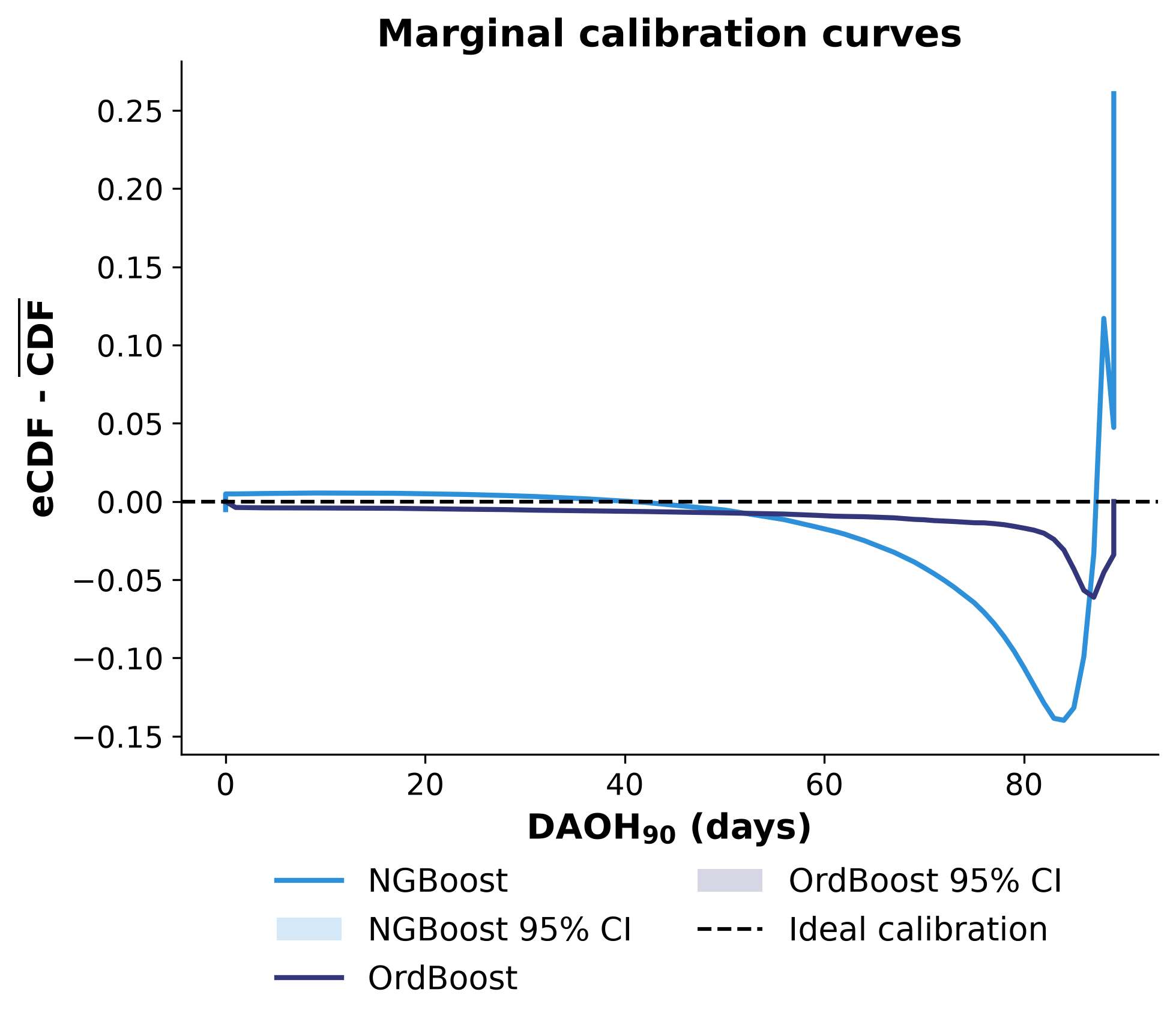}}
\end{figure}(above the ideal line), indicating that the prediction intervals were too wide and included the true $DAOH_{90}$ value too often. However, the OrdBoost model had better coverage reliability compared to the NGBoost model. The OrdBoost model was sharper than the NGBoost model (\figureref{fig:sharpness}). The mean interval width of the prediction intervals was wider for the NGBoost model due to its over-dispersion and under-confidence. In contrast, the OrdBoost model had tighter interval widths on 
average, except at 95\% nominal coverage, where the NGBoost model had smaller intervals.   

\figureref{fig:cdfs} shows the CDFs generated by the NGBoost (\figureref{fig:cdfs} panel A) and OrdBoost (\figureref{fig:cdfs} panel B) models for patients in the test set at different $DAOH_{90}$ values. The median line was plotted as a reference. For the OrdBoost model, the CDFs accumulated 50\% of their total mass near the expected $DAOH_{90}$ value, as indicated by the intersection of the CDFs with the median line.

\tableref{tab:tab2} shows the predictions made by all four models for five patients in the test set at different $DAOH_{90}$ values. For the NGBoost and OrdBoost models, the median prediction was made from the CDFs shown \begin{figure}[h]
\floatconts
  {fig:cdfs}
  {\caption{Example of CDFs from the two models for $DAOH_{90}$ values across the range.}}
  {%
    \subfigure[CDFs from the NGBoost model for patients with $DAOH_{90}$ values in $\{0, 20, 70, 85, 89\}$.]{\label{fig:cdfs_ngboost}%
      \includegraphics[width=\linewidth]{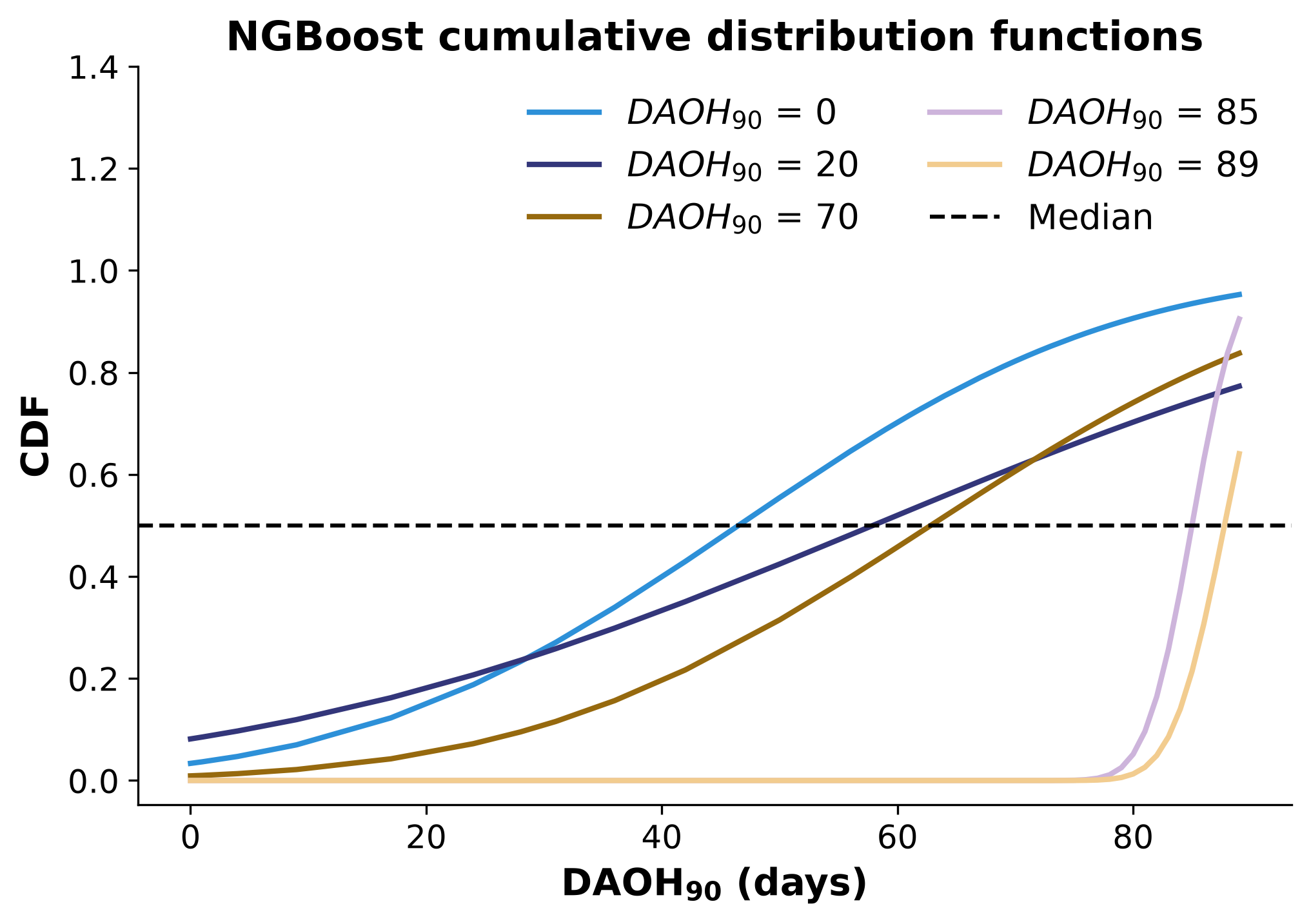}}%
    \qquad
    \subfigure[CDFs from the OrdBoost model for patients with $DAOH_{90}$ values in $\{0, 20, 70, 85, 89\}$.]{\label{fig:cdfs_ordboost}%
      \includegraphics[width=\linewidth]{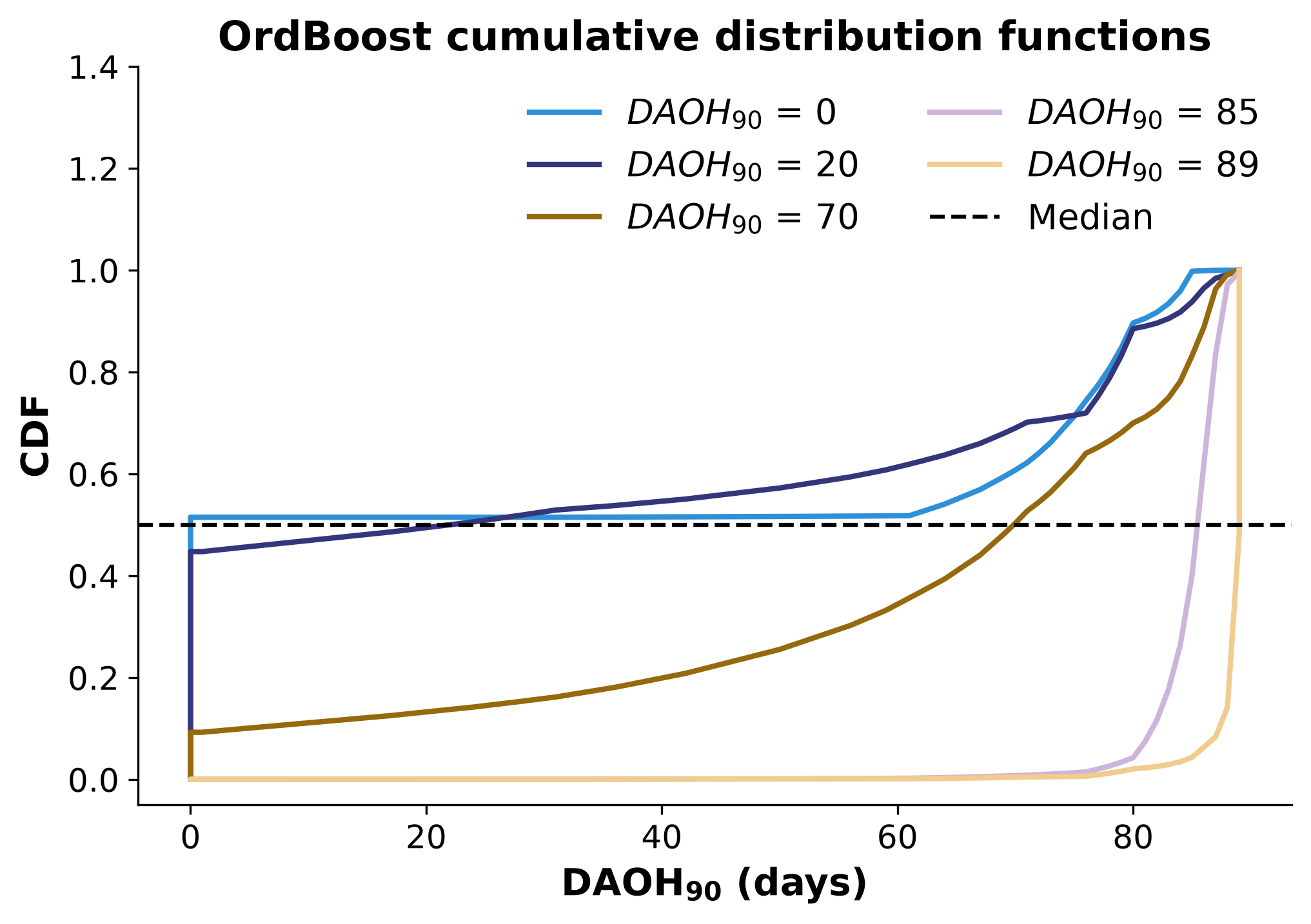}}
  }
\end{figure}\figureref{fig:cdfs} and presented along with the 95\% prediction interval (PI). The HGBRegressor and OrdinalRidge models only provided point-estimates for each patient, as they do not natively produce prediction intervals. The OrdBoost model predicted median values close to the true $DAOH_{90}$ for all five patients. On the other hand, the three other models overestimated the $DAOH_{90}$ value for the first two patients, but performed well for the three patients with $DAOH_{90} \ge 70$. For the NGBoost model, these values were consistent with the marginal calibration, which showed, on average, overestimation in the $DAOH_{90} \le 40$ range and underestimation in the $40 < DAOH_{90} \le 85$ range. The 95\% PIs, in both OrdBoost and NGBoost models, were wide for the first three patients and narrowed for the last two patients for both models.

\begin{sidewaystable}[htbp]
\floatconts
  {tab:tab2}
  {\caption{$DAOH_{90}$ predictions for all models with 95\% PI when available.}}
  {\begin{tabular}{lcccc}
\toprule
$\mathbf{DAOH_{90}}$ & \textbf{HGBRegressor} & \textbf{OrdinalRidge} & \textbf{NGBoost} & \textbf{OrdBoost} \\
days & point-estimate & point-estimate & median [95\% PI] & median [95\% PI] \\
\midrule
0 & 49 & 57 & 47 [0--89] & 0 [0--85] \\
20 & 63 & 68 & 58 [0--89] & 22 [0--87] \\
70 & 69 & 72 & 63 [11--89] & 70 [0--88] \\
85 & 85 & 87 & 85 [79--89] & 85 [78--89] \\
89 & 88 & 86 & 88 [81--89] & 89 [82--89] \\
\bottomrule
\end{tabular}
}
\end{sidewaystable}

\begin{figure}[h]
\floatconts
  {fig:pred_dist}
  {\caption{Predicted $DAOH_{90}$ distributions on the test set for all four models. Only the OrdBoost model captured the spike at 0 days.}}
  {\includegraphics[width=\linewidth]{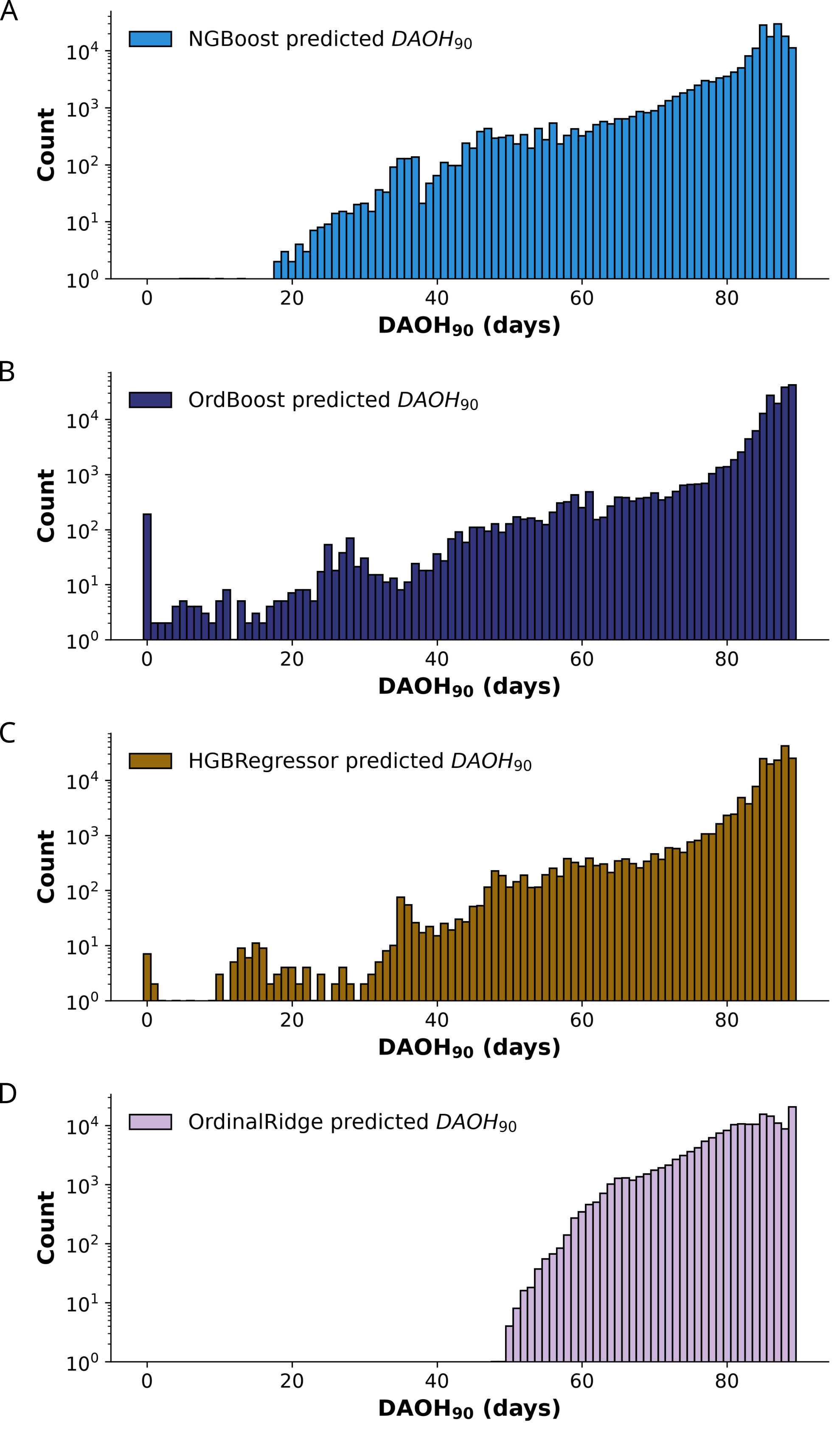}}
\end{figure}
The predicted $DAOH_{90}$ distributions are displayed in \figureref{fig:pred_dist}. All four models predicted the upper region of the distribution well; however, the NGBoost and OrdinalRidge models were unable to predict the lower values. The OrdinalRidge model produced the worst distribution, with no predicted values below 48 days. On the other hand, both HGBRegressor and OrdBoost models predicted the entire range of $DAOH_{90}$ values. However, only the OrdBoost model captured some of the zero-inflation of the data with a clear spike at 0 days.

\section{Discussion}
\label{sec:disc}
We presented a novel method using ordinal-binning gradient-boosting (OrdBoost) for predicting the patient-centred outcome of days alive and out of hospital at 90 days ($DAOH_{90}$). The target outcome was divided into ordinal bins of varying sizes, and binary classifiers were trained to predict the new targets. A mapper was then fitted to convert the predictions from the classifiers back into the original target space. Mapping the predictions allowed the OrdBoost model to produce a continuous predictive function (CDF) for each patient. The model was compared to NGBoost, another regression model capable of producing CDFs \citep{duan_ngboost_2020}, and two other regression models, a histogram gradient-boosting model and an ordinal ridge model, serving as baselines. 

\subsection{Model predictions}
The difficulty in predicting $DAOH_{90}$ lay in the data distribution \citep{holodinsky_comparing_2021}. Firstly, the data distribution is complex, with a point mass at 0 for patients who died in hospital or stayed longer than the temporal horizon. Secondly, the distribution is heavily skewed, with the 10\textsuperscript{th} percentile at 78 days (see \figureref{fig:distributions}). \citet{holodinsky_comparing_2021} compared 15 different models predicting $DAOH_{90}$ for patients with a stroke diagnosis. They found that although the machine learning models performed well on accuracy and bias, they failed to capture the specific $DAOH_{90}$ data distribution. Similar results were observed here.

Overall accuracy of the four models was evaluated using MAE, CRPS, and their associated skill scores. All models showed relatively small errors with MAE values of at most 6.19 days. However, the MAESS, which compared the MAE against a model only predicting the median value with no patient information, showed that the OrdinalRidge model was less accurate (\num{-0.233} 95\%CI [\num{-0.241}, \num{-0.224}]), and that the NGBoost model was just as accurate as the naive model with a MAESS near 0. The OrdBoost and the HGBRegressor models both had similar MAE and positive skill scores. 

When looking at the predicted distributions (see \figureref{fig:pred_dist}), all models exhibited difficulties in predicting the full $DAOH_{90}$ distribution. Two models, NGBoost and OrdinalRidge, did not capture the entire range, with values below 14 and 48 days, respectively, never being predicted. The HGBRegressor model was the closest among the comparison models, with predictions spanning the full $DAOH_{90}$ range. However, similarly to findings from \citet{holodinsky_comparing_2021}, the zero-day spike was never captured. In contrast, the OrdBoost model clearly expressed this spike. However, both models had difficulty predicting the tail ends of the distributions, with lower-than-expected counts. Nevertheless, this strongly suggested that the ordinal-binning gradient-boosting strategy employed is a viable approach to modelling complex data distributions such as $DAOH_{90}$.

\subsection{Calibration}
Both OrdBoost and NGBoost models displayed some miscalibration as evidenced by the PIT histograms (\figureref{fig:pit}) and the marginal calibration curves (\figureref{fig:cal}). The OrdBoost PIT histogram was left-skewed, associated with a biased model that systematically predicts smaller values than expected \citep{gneiting_probabilistic_2007}. This was confirmed with the marginal calibration curve, which was systematically negative. The CDFs predicted by the OrdBoost model accumulated probability faster than the empirical CDF, resulting in predictions that were lower than expected. On the other hand, the NGBoost PIT histogram was more hump-shaped, indicating an overdispersed model with overly wide average prediction intervals \citep{gneiting_probabilistic_2007}. The marginal calibration curve indicated the direction of the miscalibration. It showed that the model overestimated in the $DAOH_{90} \le 40$ and $DAOH_{90} \ge 85$ ranges, but underestimated $DAOH_{90}$ in the $40 \le DAOH _ {90} \le 85$ range. Unlike the OrdBoost model that was constrained to generate complete CDFs between 0 and 89 by design, the NGBoost model had no boundaries and constructed CDFs with $F(0^-) > 0$ and $F(89) < 1$. Consequently, the CDFs under-accumulated probability mass near the boundaries, but over-accumulated in the centre, explaining the alternating sign of the marginal calibration curve and the hump in the PIT histogram. The miscalibration was exacerbated in the upper region due to the left-skew in the $DAOH_{90}$ distribution (\figureref{fig:distributions}). 

Binary classification models often employ recalibration to correct the miscalibration of a model \citep{huang_tutorial_2020}. A similar principle can be applied to forecasting models that produce CDFs \citep{qian_regression_2025,graziani_probabilistic_2021}. There are a variety of recalibration methods that involve learning a mapping between the PIT values predicted by the model and the uniform distribution, which represents perfect calibration. Recalibrating the PIT values using an unseen recalibration set could reduce the miscalibration observed in both models, especially the systematic bias from the OrdBoost model.

\subsection{Prediction intervals}
An important feature of both the OrdBoost and NGBoost models was the ability to generate prediction intervals, thereby providing a measure of uncertainty. \citet{gneiting_probabilistic_2007} explained that a model must exhibit both sharpness and coverage to be informative. In other words, the prediction intervals must contain the true value as often as expected and must be narrow enough to provide information. The OrdBoost model produced narrow prediction intervals (PIs), with a 95\% PI of 26 days on average, with excessive coverage, with 97\% of true values falling within the 95\% PIs. The NGBoost intervals were also too wide, already capturing 90\% of true values with the 60\% PIs, and averaging 22 days for the 95\% PIs. However, \tableref{tab:tab2} highlighted that the interval widths were much wider for patients with true $DAOH_{90} \le 70$, spanning almost the entire $DAOH_{90}$ range for the OrdBoost model and beyond the range for the NGBoost model before clipping. Thus, the model had much less predictive power for patients with lower $DAOH_{90}$ values, as expected given the $DAOH_{90}$ distribution: the 10\textsuperscript{th} percentile was 78 days. Therefore, there was less data to learn from in the $DAOH_{90} \le 70$ range, which produced more uncertainty and wider intervals. Nevertheless, the models still provided more information than standard regression models, which only produced point-estimate predictions.

Generating a CDF is not the only method for quantifying uncertainty in predictions. Conformal predictions, which have been used in medical settings \citep{barriga-gallegos_enhancing_2026, vazquez_conformal_2022}, are a method of quantifying uncertainty for point-estimate predictions. However, conformal predictions typically rely on an additional dataset to learn the non-conformity measure \citep{vazquez_conformal_2022}. In contrast, models such as NGBoost or OrdBoost do not require an additional dataset, but care must be taken to ensure that the PIs are informative. Although conformal prediction would add predictive uncertainty to regression models such as the HGBRegressor or the OrdinalRidge model, this would not change the predicted distribution, and these models would still fail to capture the zero-inflation. 

\subsection{Limitations}
The current strategy proved effective in predicting $DAOH_{90}$; however, $DAOH_{90}$ varies across hospitals \citep{wells_hospital_2026,boyle_using_2023}; thus, to further validate the quality of the model, a geographic validation should be conducted to verify that the model remains performant across different spatial regions.

This study compared the new OrdBoost model with an NGBoost model. However, the NGBoost model produced CDFs that were not bounded within the $DAOH_{90}$ range. Thus, the predictions and intervals were clipped to [0--89] to make the model more comparable. This was an easily implemented solution; however, other methods could be explored to ensure values fall within the correct range, such as a logistic transformation \citep{lesaffre_logistic_2007}.  

\subsection{Conclusion}
A novel modelling strategy was developed using an ordinal-binning gradient-boosted regression approach to predict $DAOH_{90}$ from Aotearoa, New Zealand surgical hospitalisation data. The model produced a CDF for each patient from which a point estimate and prediction interval can be presented. In testing, the model showed strong performance across most patients and outperformed an NGBoost comparator model, whilst being best able to capture data spikes at 0 and 89 days in the $DAOH_{90}$ distribution.

\bibliography{ref}
\onecolumn
\newpage
\twocolumn
\appendix

\section{Effects from the number of bins}
\label{sec:n_bins}
\subsection{Methods}
This study focused on the number of bins and their effects on the performance of the OrdBoost model. Five different OrdBoost models were trained with 4, 8, 12, 16, or 20 uniformly distributed bins over the $DAOH_{90}$ range. 
\tableref{tab:tab_n_bins_bins} shows the bins used by each model.

\begin{table}[htbp]
\floatconts
  {tab:tab_n_bins_bins}
  {\caption{Bins used by each model.}}
  {\begin{tabular}{cc}
\toprule
\textbf{Model} & \textbf{Bins} \\
\midrule
\textbf{4-bins} & [0 -- 23], [24 -- 45], [46 -- 67], [68 -- 89] \\
\textbf{8-bins} & [0 -- 12], [13 -- 23], [24 -- 34], [35 -- 45], \\
 & [46 -- 56], [57 -- 67], [68 -- 78], [79 -- 89] \\
\textbf{12-bins} & [0 -- 8], [9 -- 15], [16 -- 23], [24 -- 30], \\
 & [31 -- 38], [39 -- 45], [46 -- 52], [53 -- 60], \\
 & [61 -- 67], [68 -- 75], [76 -- 82], [83 -- 89] \\
\textbf{16-bins} & [0 -- 6], [7 -- 12], [13 -- 17], [18 -- 23], \\
 & [24 -- 28], [29 -- 34], [35 -- 39], [40 -- 45], \\
 & [46 -- 51], [52 -- 56], [57 -- 62], [63 -- 67], \\
 & [68 -- 73], [74 -- 78], [79 -- 84], [85 -- 89] \\
\textbf{20-bins} & [0 -- 5], [6 -- 9], [10 -- 14], [15 -- 18], \\
 & [19 -- 23], [24 -- 27], [28 -- 32], [33 -- 36], \\
 & [37 -- 41], [42 -- 45], [46 -- 49], [50 -- 54], \\
 & [55 -- 58], [59 -- 63], [64 -- 67], [68 -- 72], \\
 & [73 -- 76], [77 -- 81], [82 -- 85], [86 -- 89] \\
\bottomrule
\end{tabular}
}
\end{table}

The models were compared based on three metrics: \begin{enumerate}
    \item the overall accuracy, measured with the MAE, CRPS and associated skill scores;
    \item the calibration, evaluated with the marginal calibration curve and the PIT histogram; and
    \item the quality of the prediction intervals, measured with the Winkler score, which accounts for the trade-off between coverage and sharpness \citep{winkler_decision-theoretic_1972}. 
\end{enumerate} 
The 95\% confidence intervals were computed via bootstrapping with 1,000 iterations.

The models were trained with the same hyperparameters as the OrdBoost model. Training MAE and CRPS were evaluated with 10-fold cross-validation and are shown in \appendixref{sec:train}. The evaluation was performed on the held out validation set from the 2023 operation year.

\subsection{Results}

\tableref{tab:n_bin_res} shows the accuracy results for the five models. Increasing the number of bins improved all four metrics, with models having more than 8 bins showing positive skill scores. The results appeared to converge as the number of bins increased.  

\begin{sidewaystable}[htbp]
\floatconts
  {tab:n_bin_res}
  {\caption{MAE, CRPS and skill scores with 95\% CI for the five models with varying numbers of bins.}}
  {\begin{tabular}{lccccc}
\toprule
\textbf{Metrics} & \textbf{4-bins} & \textbf{8-bins} & \textbf{12-bins} & \textbf{16-bins} & \textbf{20-bins} \\
\midrule
\textbf{MAE (days)} & 7.77 [7.73, 7.81] & 4.95 [4.91, 5.00] & 4.22 [4.18, 4.27] & 4.06 [4.02, 4.11] & 3.82 [3.78, 3.86] \\
\textbf{MAESS} & -0.607 [-0.621, -0.595] & -0.025 [-0.031, -0.019] & 0.126 [0.122, 0.131] & 0.159 [0.155, 0.164] & 0.209 [0.205, 0.214] \\
\textbf{CRPS (days)} & 4.99 [4.97, 5.02] & 3.51 [3.48, 3.54] & 3.15 [3.12, 3.19] & 3.02 [2.98, 3.05] & 2.95 [2.91, 2.98] \\
\textbf{CRPSS} & -0.224 [-0.234, -0.215] & 0.139 [0.134, 0.145] & 0.227 [0.222, 0.232] & 0.261 [0.256, 0.266] & 0.278 [0.273, 0.283] \\
\bottomrule
\end{tabular}
}
\end{sidewaystable}

\figureref{fig:n_bins_cal} shows the marginal calibration curve as the number of bins increases. The sparse tail region ($DAOH_{90}< 70$) was not as affected by the increasing number of bins as the denser upper region. As the number of bins increased and covered more of the upper values, the miscalibration spike shifted and peaked near 85 days.

The PIT histograms, displayed in \figureref{fig:n_bins_pit}, became more uniform as the number of bins increased. With more bins in the upper, denser region, the PIT values were distributed more evenly, reducing the large spike at 0.75 present in the 4-bin model. The $\alpha$ score decreased with the number of bins as the histograms tended towards the ideal uniform distribution. 
\begin{figure}[h]
\floatconts
  {fig:n_bins_cal}
  {\caption{Marginal calibration curves for five models with a varying number of bins. As the number of bins increased, the model miscalibration in the upper region grew. The shaded area representing the 95\% CI reduced in size as the number of bins increased.}}
  {\includegraphics[width=\linewidth]{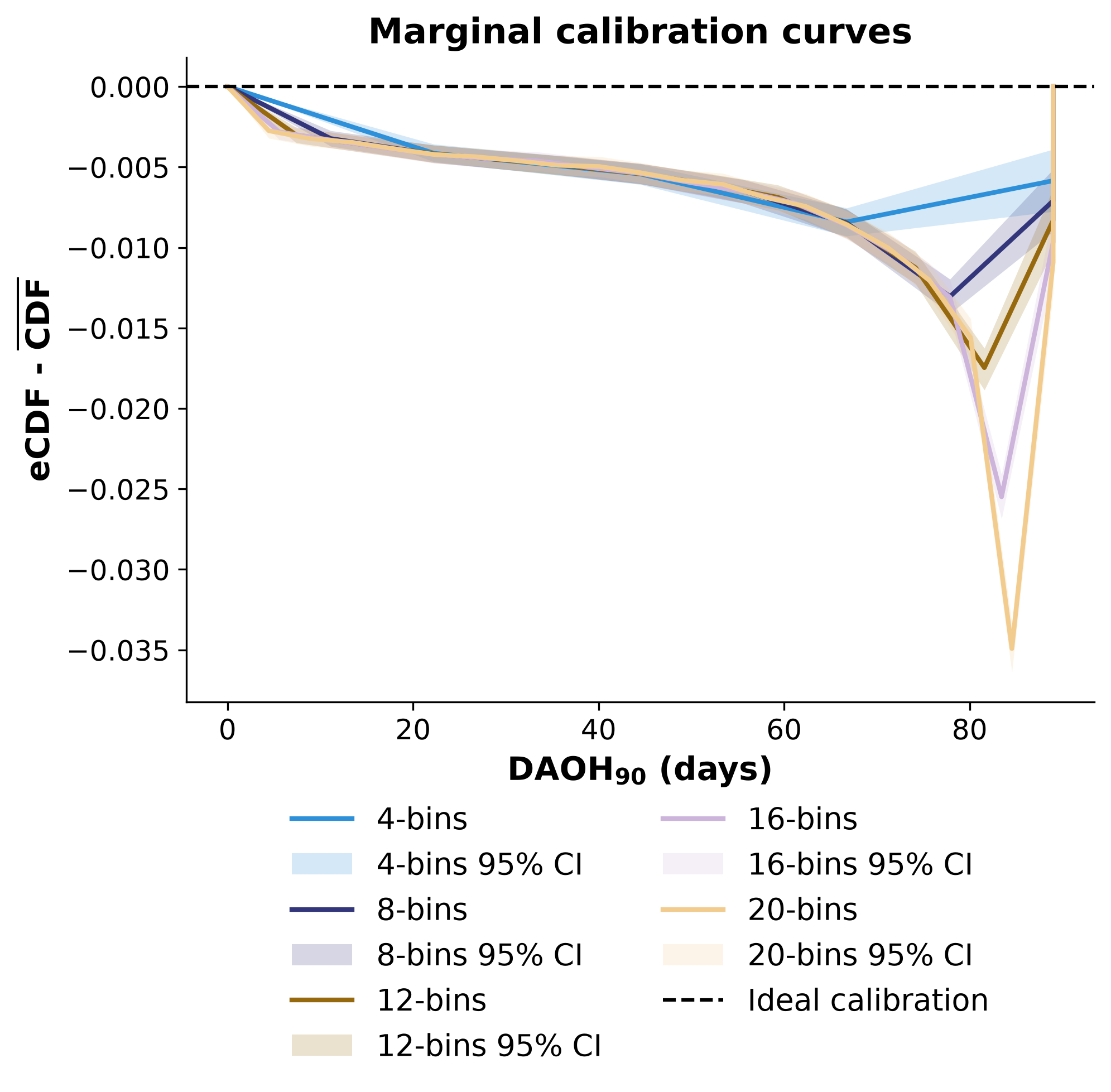}}
\end{figure}
\begin{figure}[h]
\floatconts
  {fig:n_bins_pit}
  {\caption{PIT histograms for five models with a varying number of bins. The PIT histograms became more uniform as the number of bins increased, with the spike near 0.75 being dispersed. The error bars represented the 95\% CI.}}
  {\includegraphics[width=\linewidth]{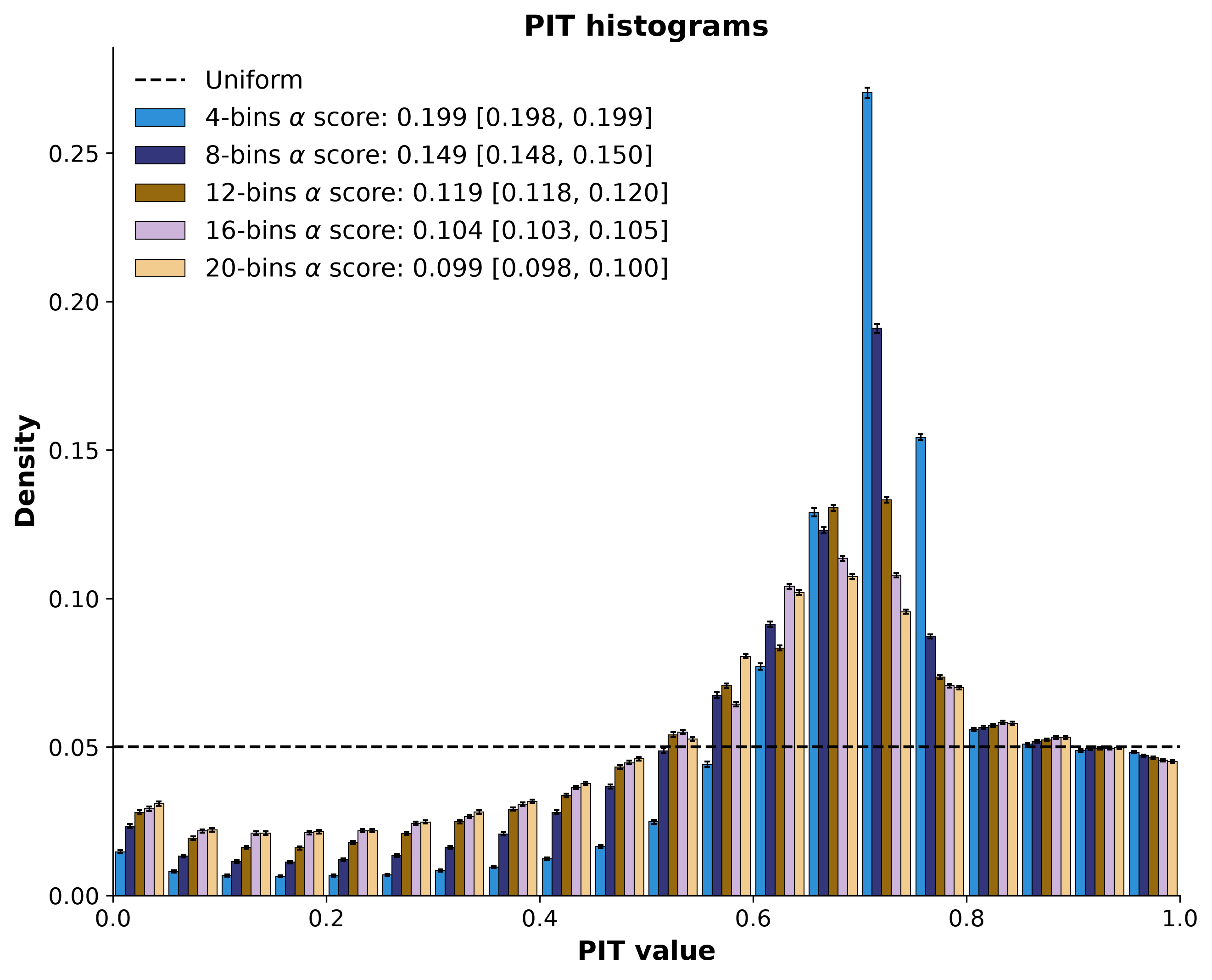}}
\end{figure}

The Winkler scores assessed the quality of the prediction intervals and captured the trade-off between sharpness and coverage. \figureref{fig:n_bins_winkler} shows the scores at different nominal coverage levels for the five models. As the number of bins increased, the scores converged, with a Winkler score near 33 at 95\% nominal coverage. 

\begin{figure}[h]
\floatconts
  {fig:n_bins_winkler}
  {\caption{The Winkler scores at different nominal coverage levels for five models with a varying number of bins. The scores converged as the number of bins increased. Error bars represented 95\% CI.}}
  {\includegraphics[width=\linewidth]{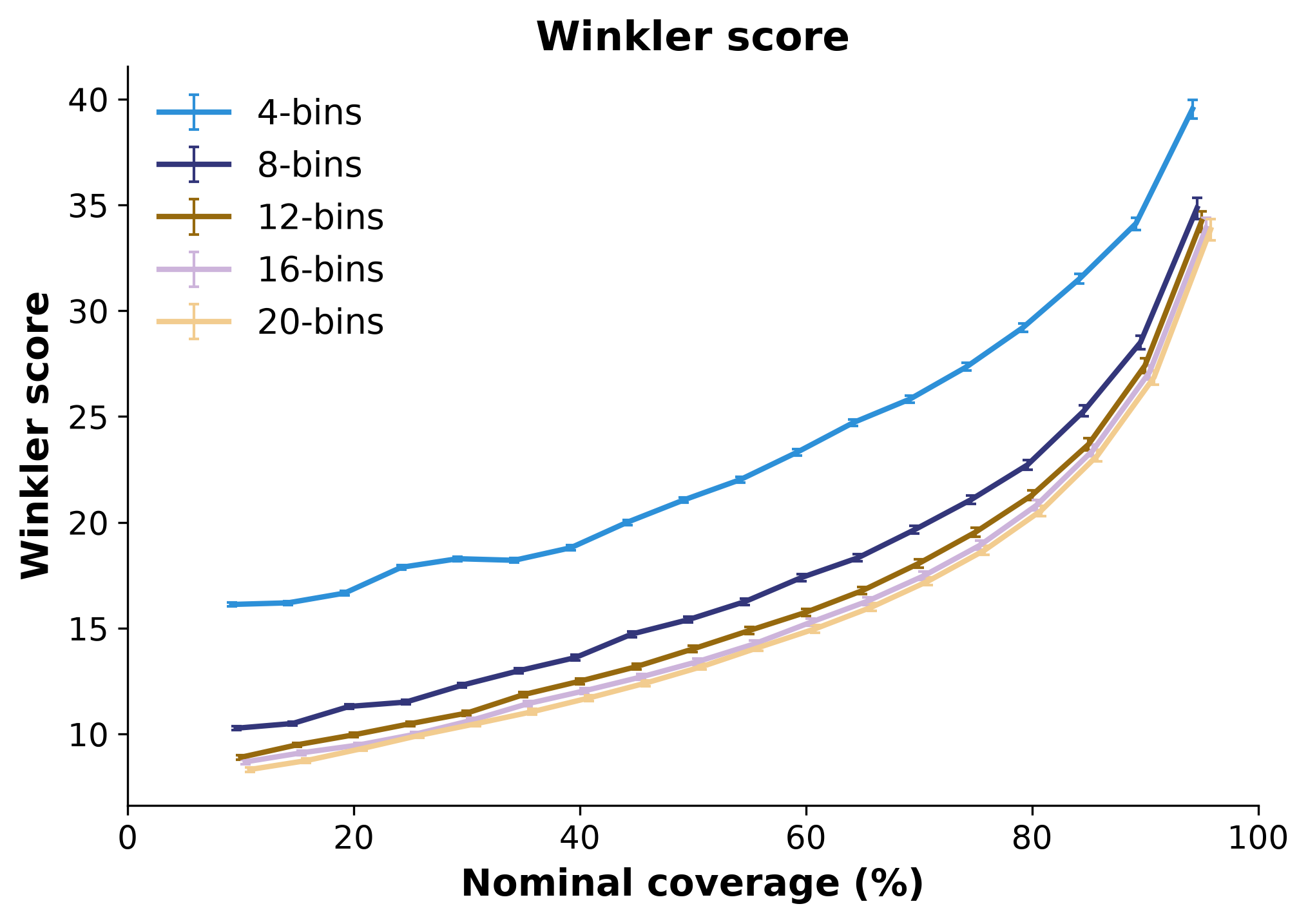}}
\end{figure}

\subsection{Conclusion}
The results showed that the number of bins strongly influenced accuracy, calibration, and prediction intervals. As the number of bins increased, the results converged, showing smaller performance gains between 16 and 20 bins than between 8 and 16 bins. Notably, 16 and 20 bins reduced the CRPS by 0.13 and 0.07 days, respectively, the $\alpha$ score decreased only from 0.119 to 0.104 and 0.099, and the Winkler scores were nearly identical. A larger number of bins requires more classifiers to fit, increasing fitting times and creating a trade-off between computation and performance. Other factors are likely to influence results, for example, a binning strategy that accounts for the zero-inflated and left-skewed distribution. Additional bins added with a uniform strategy increase the resolution in the sparse tail rather than in the denser upper region. The increase in the miscalibration near 85 days with the number of bins suggested that the placement of the bins may be driving the remaining error. As a result, 12 bins were selected and used for the next ablation studies.

\section{Effects from the bin distribution}
\label{sec:bin_dist}
\subsection{Methods}
Based on the results from the study on the number of bins (see \appendixref{sec:n_bins}), 12 bins were distributed in this study. Three different strategies were used to distribute the bins across the $DAOH_{90}$ range. The first was the previously employed uniform strategy, in which the 12 bins were distributed uniformly across the [0--89] range. The second strategy used the $DAOH_{90}$ distribution to place the bins. The 12 bin edges were selected using 13 quantiles linearly distributed along [0, 1]. The quantiles were [0, 0.08, 0.17, 0.25, 0.33, 0.42, 0.5, 0.58, 0.67, 0.75, 0.83, 0.92, 1]; however, because duplicates were removed, the number of bins was only eight (see \tableref{tab:tab_n_bin_bin_dist}). The final strategy was custom, with bins sparsely distributed in the left tail of the $DAOH_{90}$ distribution and more densely allocated in the denser region. 

The models were trained with the same hyperparameters as the OrdBoost model. Training MAE and CRPS were evaluated using 10-fold cross-validation, and the results are presented in \appendixref{sec:train}. The evaluation process was similar to the previous study (see \appendixref{sec:n_bins}). 

\begin{table}[htbp]
\floatconts
  {tab:tab_n_bin_bin_dist}
  {\caption{Bins used by each model.}}
  {\begin{tabular}{cc}
\toprule
\textbf{Model} & \textbf{Bins} \\
\midrule
Uniform & [0 -- 8], [9 -- 15], [16 -- 23], [24 -- 30], \\
 & [31 -- 38], [39 -- 45], [46 -- 52], [53 -- 60], \\
 & [61 -- 67], [68 -- 75], [76 -- 82], [83 -- 89] \\
Quantile & [0 -- 75], [76 -- 83], [84], [85], \\
 & [86], [87], [88], [89] \\
Custom & [0], [1 -- 30], [31 -- 60], [61 -- 70], \\
 & [71 -- 75], [76 -- 79], [80 -- 84], [85], \\
 & [86], [87], [88], [89] \\
\bottomrule
\end{tabular}
}
\end{table}

\subsection{Results}

\begin{sidewaystable}[htbp]
\floatconts
  {tab:bin_dist_res}
  {\caption{MAE, CRPS and skill scores with 95\% CI for the three models with different bin distribution strategies.}}
  { \begin{tabular}{lccc}
\toprule
\textbf{Metrics} & \textbf{Uniform} & \textbf{Quantile} & \textbf{Custom} \\
\midrule
\textbf{MAE (days)} & 4.22 [4.18, 4.27] & 3.80 [3.75, 3.84] & 3.67 [3.63, 3.72] \\
\textbf{MAESS} & 0.126 [0.122, 0.130] & 0.214 [0.209, 0.220] & 0.240 [0.236, 0.245] \\
\textbf{CRPS (days)} & 3.15 [3.12, 3.19] & 2.95 [2.92, 2.98] & 2.80 [2.77, 2.84] \\
\textbf{CRPSS} & 0.227 [0.222, 0.232] & 0.276 [0.271, 0.282] & 0.313 [0.308, 0.317] \\
\bottomrule
\end{tabular}
}
\end{sidewaystable}
\tableref{tab:bin_dist_res} shows the MAE, CRPS, MAESS, and CRPSS for all three models with different bin distribution strategies. All strategies proved more accurate than the naive model, with positive MAESS and CRPSS values. The Uniform strategy was the least accurate, with a MAE of 4.22 days (95\% CI [4.18, 4.27]) and a CRPS of 3.15 days (95\% CI [3.12, 3.19]). Both the Quantile and Custom strategies outperformed or matched the\begin{figure}[h]
\floatconts
  {fig:bin_dist_cal}
  {\caption{Marginal calibration curves for the three models with different bin distributions. Both the Quantile and the Custom strategies exhibited similar miscalibration to the 20-bin model from the previous study.}}
  {\includegraphics[width=\linewidth]{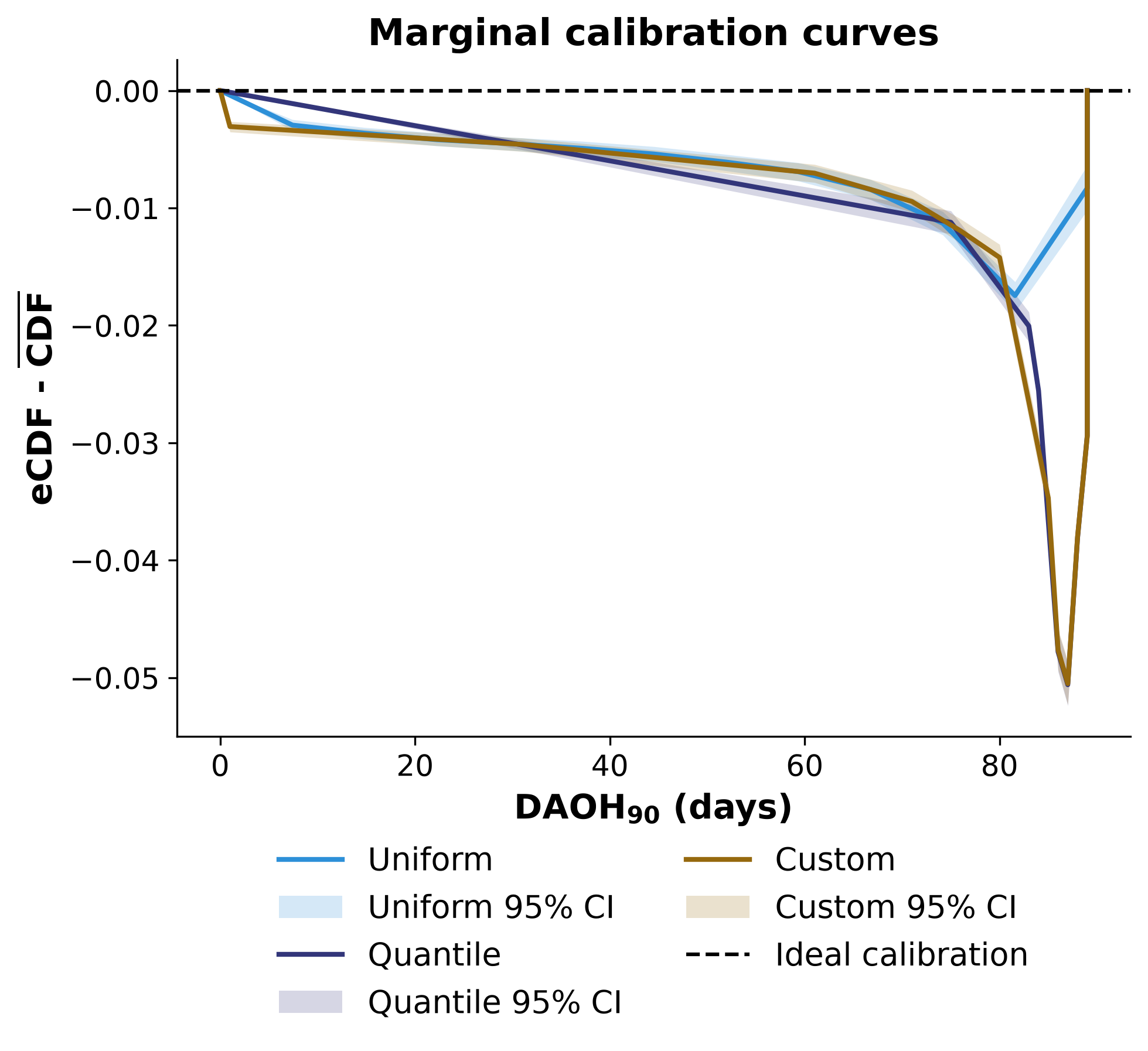}}
\end{figure} 20-bin model with a Uniform strategy. However, the Custom strategy had the best results, with a MAE of 3.67 days (95\% CI [3.63, 3.72]) and a CRPS of 2.80 days (95\% CI [2.77, 2.84]).

\figureref{fig:bin_dist_cal} shows the marginal calibration curve for the three distribution strategies. The calibration in the left tail was better for the Uniform and the Custom strategies because the bins covered that region more finely. The Quantile and Custom strategies exhibited the same miscalibration spike near 85 days as the uniform strategy with 20 bins.

The PIT histograms are displayed in \figureref{fig:bin_dist_pit}. They showed significant improvements for the Quantile and Custom strategies over the Uniform strategy. The $\alpha$ scores were approximately halved from 0.119 to 0.054 and 0.053 for the Quantile and Custom strategies, respectively. The PIT histograms for these two strategies were left-skewed, indicating a systematic bias towards underestimating the target.

\begin{figure}[h]
\floatconts
  {fig:bin_dist_pit}
  {\caption{PIT histograms for the three models with different bin distributions. The PIT histograms for the Quantile and Custom strategies were more uniform than those of the Uniform strategy. The $\alpha$ scores were lower than those of the model with 20 bins (\figureref{fig:n_bins_pit}). Error bars represented the 95\% CI.}}
  {\includegraphics[width=\linewidth]{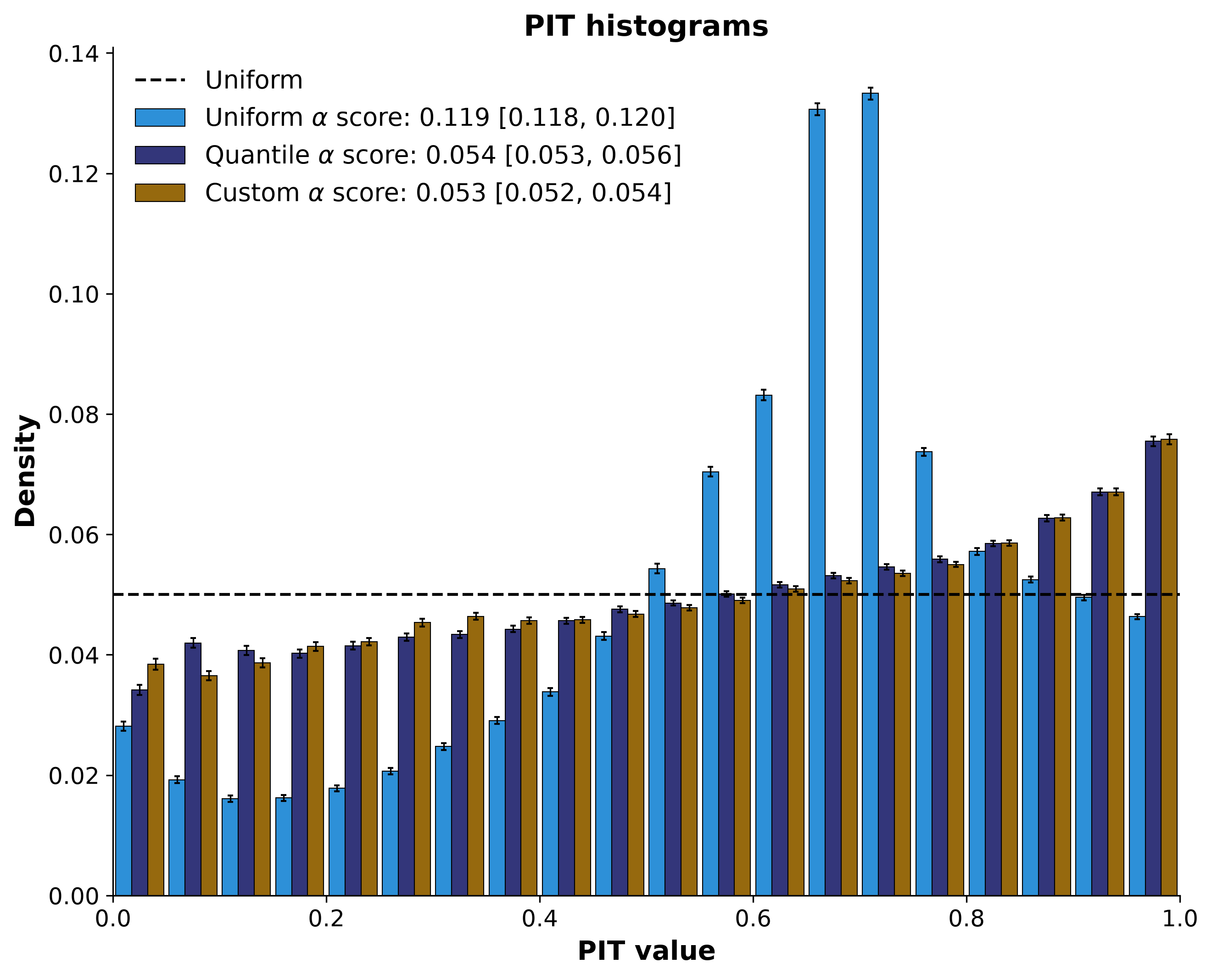}}
\end{figure}
\begin{figure}[h]
\floatconts
  {fig:bin_dist_winkler}
  {\caption{The Winkler scores at different nominal coverage rates for the three models with different bin distribution strategies. The scores clearly showed that the Custom strategy had better prediction intervals than the two other models. Error bars represented 95\% CI.}}
  {\includegraphics[width=\linewidth]{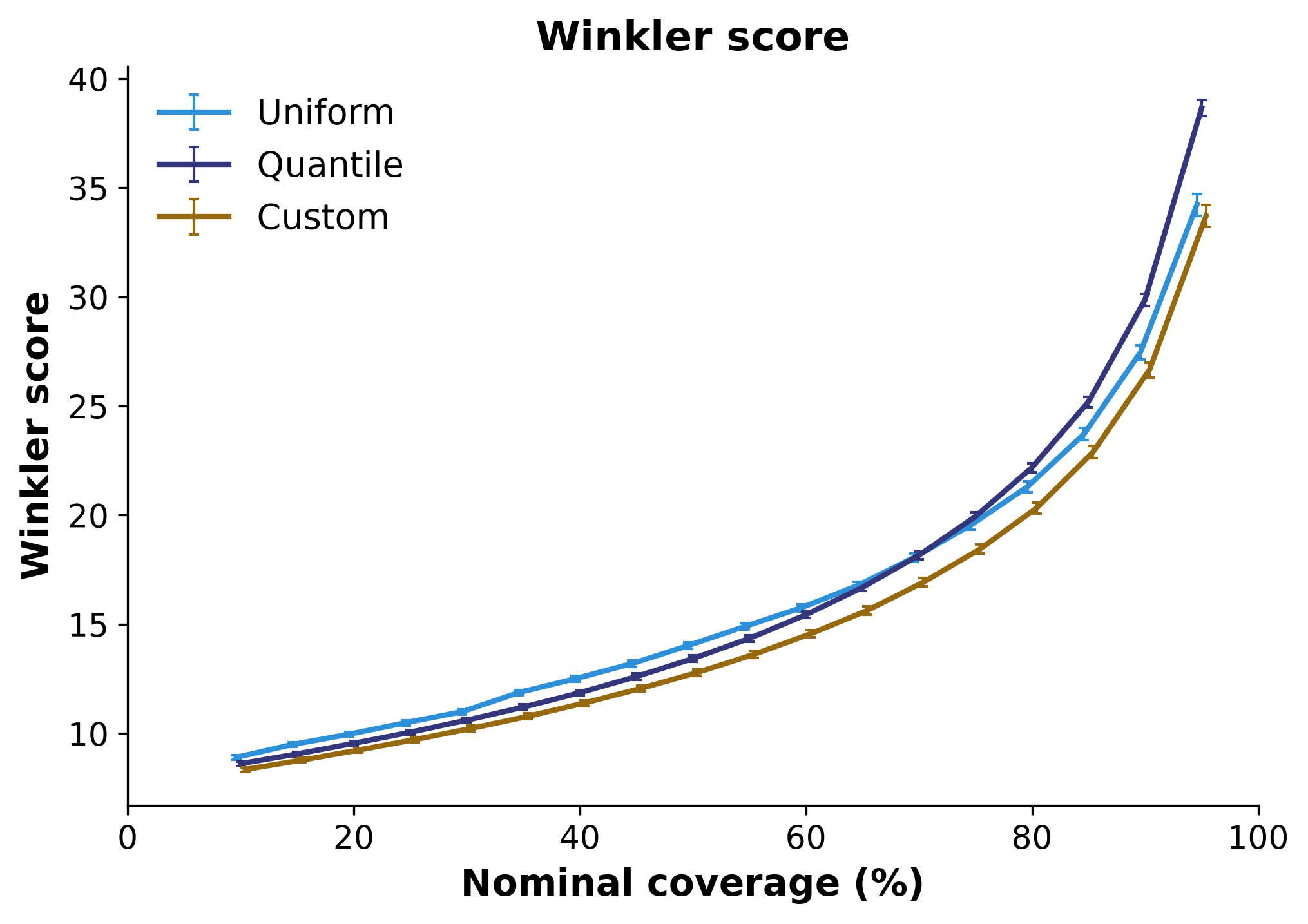}}
\end{figure}

\figureref{fig:bin_dist_winkler} shows the scores at different nominal coverage rates for the three models with different bin distribution strategies. The Custom strategy clearly outperformed both the Uniform and Quantile strategies. Although the Quantile model was well calibrated, the Winkler score was worse than that of the Uniform strategy, indicating poorer-quality prediction intervals.

\subsection{Conclusion}
With an equal number of bins, the distribution strategy clearly influenced the performance of the model. Both the Quantile and Custom strategies accounted for the $DAOH_{90}$ distribution, allowing the models to match and outperform those with 16 and 20 bins in the previous study. The Custom strategy outperformed the other two strategies across all three metrics, as well as the 20-bin model from the previous study (\appendixref{sec:n_bins}). This further justified the selection of the number of bins. Thus, the Custom strategy bins were employed for the main model and the subsequent study involving mapping strategies. 

\section{Effects from the mapping strategy}
\label{sec:mappers}
\subsection{Methods}
The bin mapper was the component of the OrdBoost model that mapped predictions from discrete bins back to days. Five different mapping strategies were evaluated: 
\begin{enumerate}
    \item the uniform mapper, which provided no within-bin information to the CDF and only used the bin edges;
    \item the mean mapper, which included for each non-single-width bin, the intra-bin mean from the training data;
    \item the median mapper, which operated similarly to the mean mapper using the intra-bin median instead;
    \item the quantile mapper, which used intra-bin quantiles at [0, 0.1, 0.25, 0.5, 0.75, 0.9, 1] to add evaluation points to the CDF; and
    \item the continuous mapper, which included all possible values from 0 to 89 in the CDF.
\end{enumerate}

Based on the previous results from \appendixref{sec:n_bins} and \appendixref{sec:bin_dist}, 12 bins distributed according to the custom strategy in \tableref{tab:tab_n_bin_bin_dist} were used for all five models. The models were trained with the same hyperparameters as the OrdBoost model. Training MAE and CRPS were evaluated with 10-fold cross-validation. The evaluation process was similar to the previous study (see \appendixref{sec:n_bins}).  

\subsection{Results}
\tableref{tab:mappers_res} shows the MAE, CRPS, MAESS, and CRPSS for the five models with different mappers. The results were near-identical with overlapping CIs across all mappers, indicating that the mapping strategy did not affect the accuracy of the model. 

\begin{sidewaystable}[htbp]
\floatconts
  {tab:mappers_res}
  {\caption{MAE, CRPS and skill scores with 95\% CI for the five models with different mappers.}}
  { \begin{tabular}{lccccc}
\toprule
\textbf{Metrics} & \textbf{Mean} & \textbf{Median} & \textbf{Uniform} & \textbf{Continuous} & \textbf{Quantile} \\
\midrule
\textbf{MAE (days)} & 3.67 [3.62, 3.71] & 3.66 [3.61, 3.70] & 3.67 [3.63, 3.72] & 3.66 [3.61, 3.70] & 3.66 [3.61, 3.70] \\
\textbf{MAESS} & 0.242 [0.237, 0.246] & 0.243 [0.239, 0.247] & 0.240 [0.236, 0.245] & 0.243 [0.239, 0.247] & 0.243 [0.239, 0.247] \\
\textbf{CRPS (days)} & 2.80 [2.76, 2.83] & 2.79 [2.76, 2.82] & 2.80 [2.77, 2.84] & 2.79 [2.76, 2.82] & 2.79 [2.76, 2.82] \\
\textbf{CRPSS} & 0.315 [0.310, 0.319] & 0.315 [0.311, 0.320] & 0.313 [0.308, 0.317] & 0.316 [0.312, 0.320] & 0.316 [0.312, 0.321] \\
\bottomrule
\end{tabular}
}
\end{sidewaystable}

\figureref{fig:mappers_cal} shows the marginal calibration curve for the five mapping strategies. The curves showed very little difference, with the Uniform mapping strategy exhibiting slightly greater miscalibration than the other four methods.

Similarly to the marginal calibration curves, the PIT histograms plotted in \figureref{fig:mappers_pit} were comparable. \begin{figure}[h]
\floatconts
  {fig:mappers_cal}
  {\caption{Marginal calibration curves for five models with different mapping strategies. The mapper had very little effect on the marginal calibration curves.}}
  {\includegraphics[width=\linewidth]{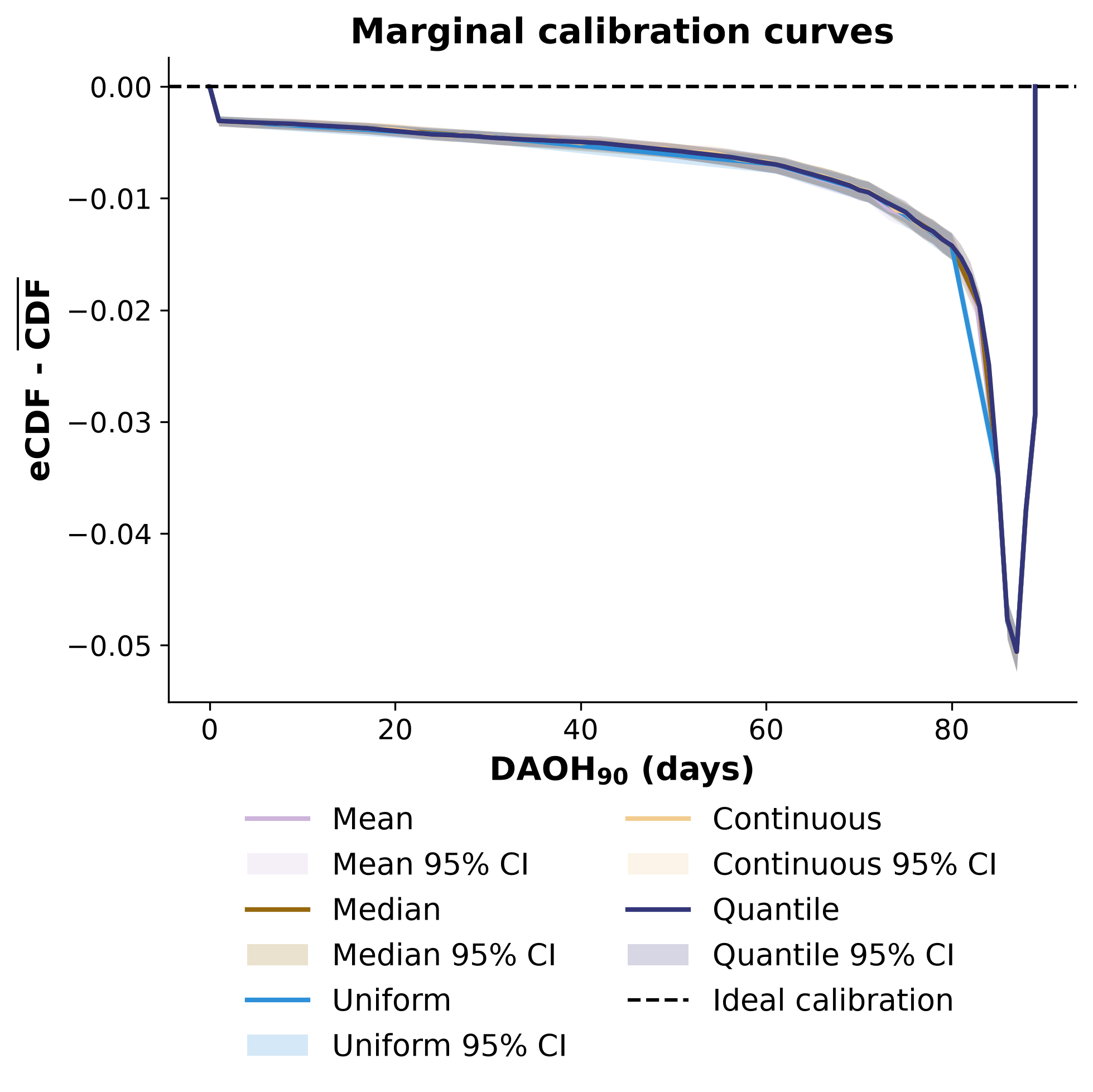}}
\end{figure}
\begin{figure}[h]
\floatconts
  {fig:mappers_pit}
  {\caption{PIT histograms for five models with different mapping strategies. The PIT histograms were all nearly uniform, with very low $\alpha$ scores. The Uniform strategy, however, did not perform as well as the four other mappers. The error bars represented the 95\% CI.}}
  {\includegraphics[width=\linewidth]{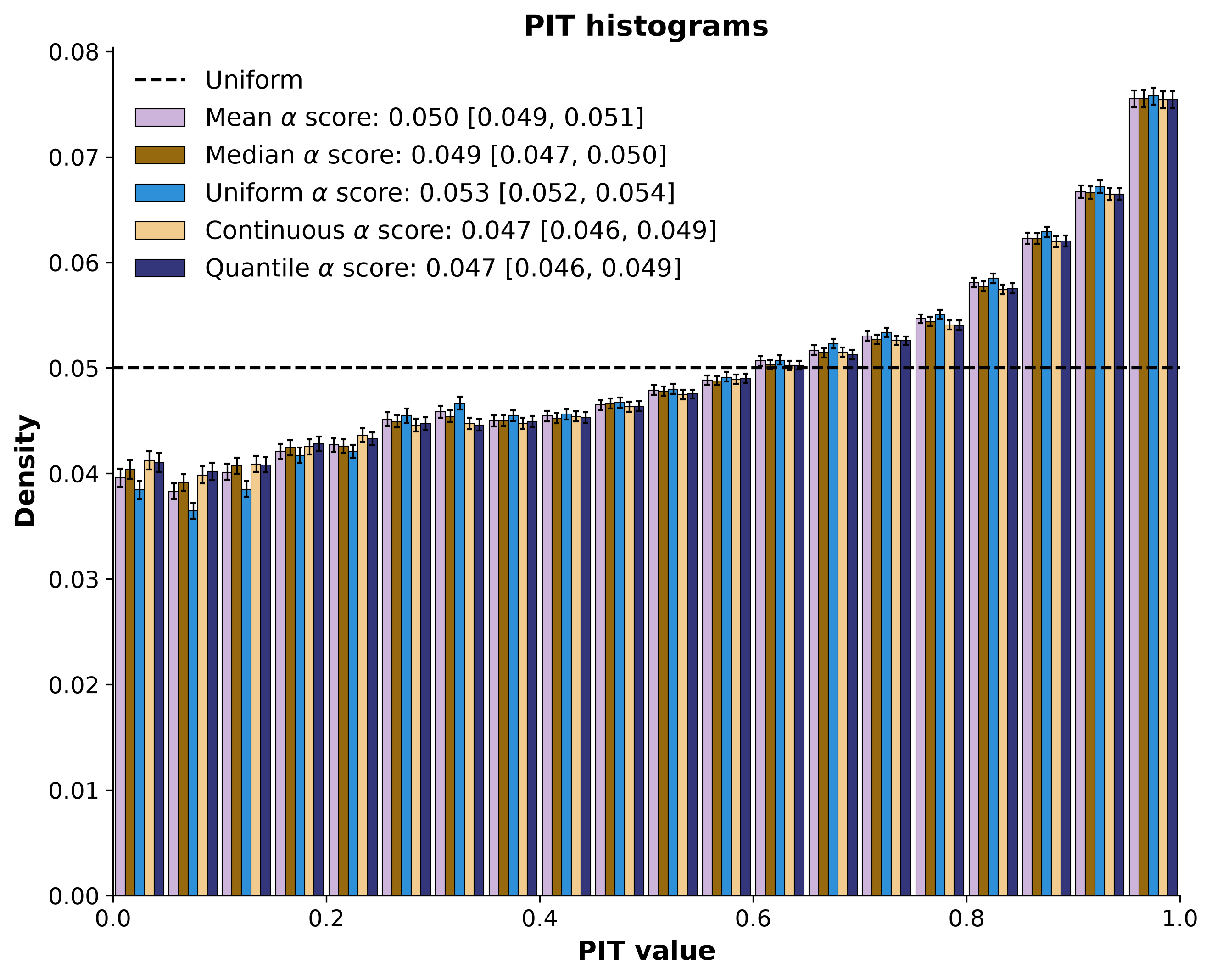}}
\end{figure}
The Uniform strategy had a higher $\alpha$ score at 0.053. The Continuous, Mean, Median, and Quantile strategies had the same $\alpha$ score given the 95\% confidence intervals. 

\figureref{fig:mappers_winkler} shows the Winkler scores at different coverage levels for the five models. The scores were identical across all models and were similar to those observed in the previous study.  

\begin{figure}[h!]
\floatconts
  {fig:mappers_winkler}
  {\caption{The Winkler scores at different nominal coverage levels for the five models with different mappers. The scores showed no differences. Error bars represented 95\% CI.}}
  {\includegraphics[width=\linewidth]{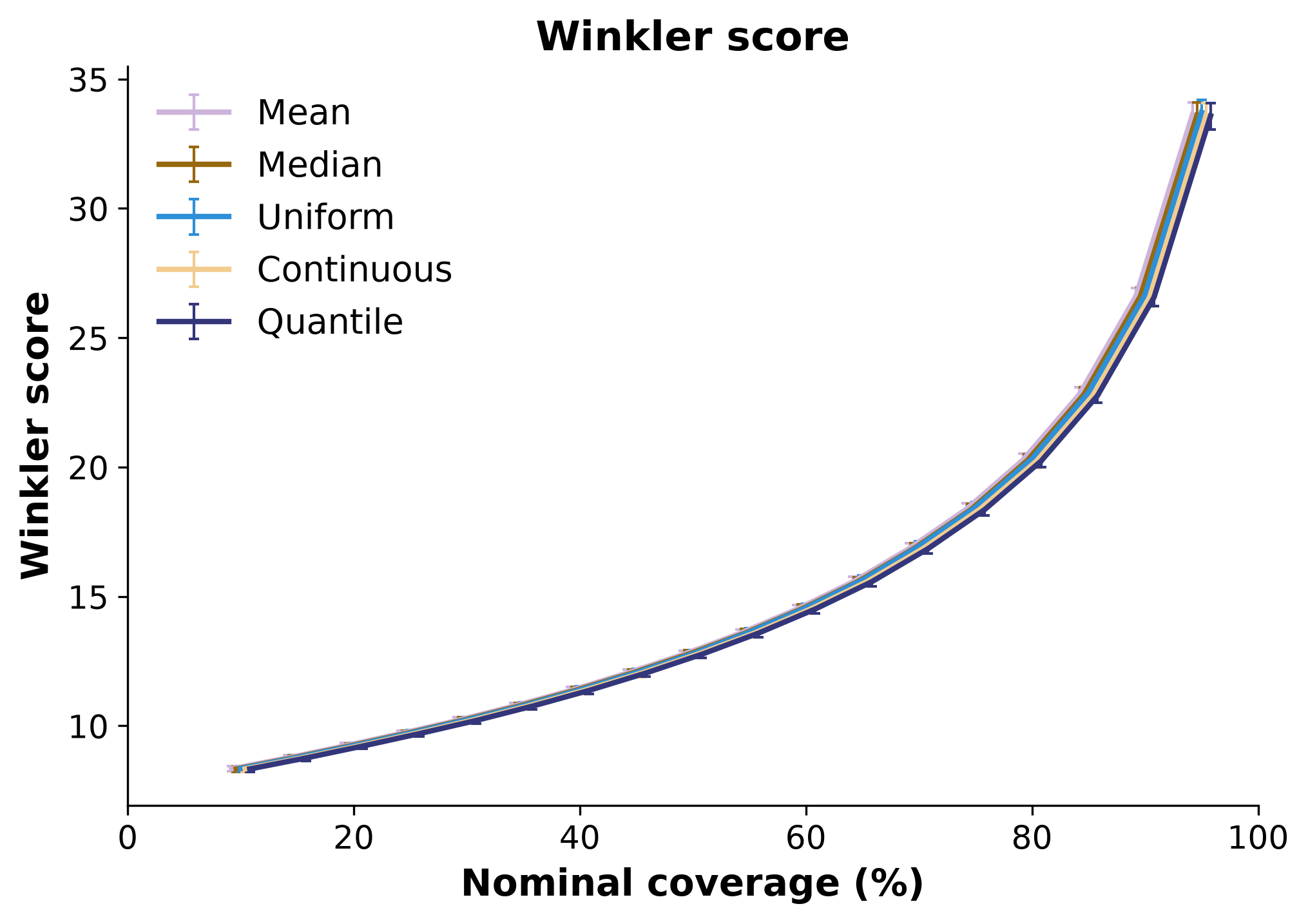}}
\end{figure}

\subsection{Conclusion}
The mapping strategy was the component that had the least effect on overall performance. It primarily affected the calibration, with PIT histograms showing marginally higher $\alpha$ scores for the Uniform mapper and marginal calibration curves closer to the ideal calibration line for the Quantile strategy. The Winkler scores and CRPS were identical across mapping strategies; thus, the Quantile mapper was preferred because it had marginally better calibration performances.

\begin{table}[!tbp]
\floatconts
  {tab:tab_training}
  {\caption{Training MAE and CRPS as mean $\pm$ std from 10-fold cross-validation.}}
  {\begin{tabular}{ccc}
\toprule
\textbf{Models}      & \textbf{MAE (days) }     & \textbf{CRPS (days)}     \\ \midrule
\textbf{HGBRegressor} & $3.65 \pm 0.02$ & $3.65 \pm 0.02$ \\ 
\textbf{OrdinalRidge} & $5.43 \pm 0.02$ & $5.43 \pm 0.02$ \\ 
\textbf{NGBoost}      & $4.43 \pm 0.01$ & $3.47 \pm 0.01$ \\ 
\textbf{OrdBoost}     & $3.70 \pm 0.02$ & $2.79 \pm 0.01$ \\ \bottomrule
\end{tabular}
}
\end{table}

\begin{table}[!tbp]
\floatconts
  {tab:tab_n_bins_train}
  {\caption{Training MAE and CRPS as mean $\pm$ std from 10-fold cross-validation for the bin number study.}}
  {\begin{tabular}{ccc}
  \toprule
\textbf{Models} & \textbf{MAE (days)}      & \textbf{CRPS (days)}     \\ \midrule
\textbf{4-bins}  & $7.52 \pm 0.01$ & $4.96 \pm 0.01$ \\
\textbf{8-bins}  & $4.84 \pm 0.01$ & $3.49 \pm 0.01$ \\
\textbf{12-bins} & $4.27 \pm 0.01$ & $3.15 \pm 0.01$ \\
\textbf{16-bins} & $4.02 \pm 0.02$ & $3.01 \pm 0.01$ \\
\textbf{20-bins} & $3.87 \pm 0.01$ & $2.93 \pm 0.01$ \\
\bottomrule
\end{tabular}
}
\end{table}

\begin{table}[!tbp]
\floatconts
  {tab:tab_bin_dist_train}
  {\caption{Training MAE and CRPS as mean $\pm$ std from 10-fold cross-validation for the bin distribution study.}}
  {
\begin{tabular}{ccc}
\toprule
\textbf{Models}  & \textbf{MAE (days)} & \textbf{CRPS (days)} \\ \midrule
\textbf{Uniform}  & $4.27 \pm 0.02$ & $3.15 \pm 0.01$ \\
\textbf{Quantile} & $3.78 \pm 0.01$ & $2.89 \pm 0.01$ \\
\textbf{Custom}   & $3.71 \pm 0.02$ & $2.80 \pm 0.01$ \\
\bottomrule
\end{tabular}
}
\end{table}

\begin{table}[!tbp]
\floatconts
  {tab:tab_mappers_train}
  {\caption{Training MAE and CRPS as mean $\pm$ std from 10-fold cross-validation for the mapping strategy study.}}
  {
\begin{tabular}{ccc}
\toprule
\textbf{Models}    & \textbf{MAE (days)} & \textbf{CRPS (days)} \\ \midrule
\textbf{Mean}       & $3.71 \pm 0.02$ & $2.80 \pm 0.01$ \\
\textbf{Median}     & $3.71 \pm 0.02$ & $2.80 \pm 0.01$ \\
\textbf{Uniform}    & $3.71 \pm 0.02$ & $2.80 \pm 0.01$ \\
\textbf{Continuous} & $3.71 \pm 0.02$ & $2.80 \pm 0.01$ \\
\textbf{Quantile}   & $3.71 \pm 0.02$ & $2.80 \pm 0.01$ \\
\bottomrule
\end{tabular}
}
\end{table}
\section{Training results}
\label{sec:train}
This appendix presents the training results from 10-fold cross-validation for the main models (\tableref{tab:tab_training}) and all ablation studies (\tableref{tab:tab_n_bins_train}, \tableref{tab:tab_bin_dist_train}, and \tableref{tab:tab_mappers_train}). In all cases, the standard deviation was very low. The differences between the training and testing set were relatively small and most likely due to the temporal shift between the two sets.

\onecolumn
\section{Data characteristics}
\label{sec:data}

\begin{longtable}{l l l l}
\caption{Patient characteristics by dataset split.} \label{tab:patient_chars} \\
\toprule
\textbf{Features} & \shortstack[l]{\textbf{Train (2010--2022)} \\ N = 2,239,441} & \shortstack[l]{\textbf{Validation (2023)} \\ N = 177,395} & \shortstack[l]{\textbf{Test (2024)} \\ N = 168,750} \\
\midrule\midrule
\endfirsthead
\multicolumn{4}{l}{\textit{(continued)}} \\
\toprule
\textbf{Features} & \shortstack[l]{\textbf{Train (2010--2022)} \\ N = 2,239,441} & \shortstack[l]{\textbf{Validation (2023)} \\ N = 177,395} & \shortstack[l]{\textbf{Test (2024)} \\ N = 168,750} \\
\midrule\midrule
\endhead
\midrule
\multicolumn{4}{r}{\textit{Continued on next page}} \\
\endfoot
\bottomrule
\endlastfoot
Age, median (IQR) & 48.0 (32.0--65.0) & 50.0 (33.0--67.0) & 51.0 (34.0--68.0) \\
\midrule
\multicolumn{4}{l}{\textbf{Sex at birth}} \\
F & 1,309,873 (58.5\%) & 104,172 (58.7\%) & 103,145 (61.1\%) \\
M & 929,552 (41.5\%) & 73,203 (41.3\%) & 65,580 (38.9\%) \\
Missing & 16 (0.0\%) & 20 (0.0\%) & 25 (0.0\%) \\
\midrule
\multicolumn{4}{l}{\textbf{Ethnicity}} \\
European & 1,577,936 (70.5\%) & 114,825 (64.7\%) & 105,090 (62.3\%) \\
Asian & 180,924 (8.1\%) & 18,948 (10.7\%) & 20,597 (12.2\%) \\
Pacific Peoples & 136,394 (6.1\%) & 11,992 (6.8\%) & 11,749 (7.0\%) \\
Māori & 307,575 (13.7\%) & 27,930 (15.7\%) & 27,620 (16.4\%) \\
MELAA/Other & 34,235 (1.5\%) & 3,479 (2.0\%) & 3,442 (2.0\%) \\
Missing & 2,377 (0.1\%) & 221 (0.1\%) & 252 (0.1\%) \\
\midrule
\multicolumn{4}{l}{\textbf{Admission source}} \\
Routine & 2,190,134 (97.8\%) & 173,666 (97.9\%) & 164,776 (97.6\%) \\
Transfer & 49,307 (2.2\%) & 3,729 (2.1\%) & 3,974 (2.4\%) \\
\midrule
\multicolumn{4}{l}{\textbf{Admission acuity}} \\
Acute & 603,225 (26.9\%) & 51,616 (29.1\%) & 53,809 (31.9\%) \\
Elective & 1,636,216 (73.1\%) & 125,779 (70.9\%) & 114,941 (68.1\%) \\
\midrule
\multicolumn{4}{l}{\textbf{Trauma}} \\
False & 2,018,562 (90.1\%) & 159,654 (90.0\%) & 152,104 (90.1\%) \\
True & 220,879 (9.9\%) & 17,741 (10.0\%) & 16,646 (9.9\%) \\
\midrule
\multicolumn{4}{l}{\textbf{Prior cancer}} \\
False & 1,854,666 (82.8\%) & 142,073 (80.1\%) & 133,432 (79.1\%) \\
True & 384,775 (17.2\%) & 35,322 (19.9\%) & 35,318 (20.9\%) \\
\midrule
\multicolumn{4}{l}{\textbf{ASA score}} \\
1.0 & 521,042 (23.3\%) & 26,711 (15.1\%) & 26,352 (15.6\%) \\
2.0 & 832,621 (37.2\%) & 75,276 (42.4\%) & 80,740 (47.8\%) \\
3.0 & 342,813 (15.3\%) & 35,364 (19.9\%) & 38,673 (22.9\%) \\
4.0 & 54,880 (2.5\%) & 5,142 (2.9\%) & 5,109 (3.0\%) \\
5.0 & 1,866 (0.1\%) & 191 (0.1\%) & 197 (0.1\%) \\
Missing & 486,219 (21.7\%) & 34,711 (19.6\%) & 17,679 (10.5\%) \\
\midrule
\multicolumn{4}{l}{\textbf{Operation severity}} \\
1 & 122,089 (5.5\%) & 9,940 (5.6\%) & 8,294 (4.9\%) \\
2 & 1,034,729 (46.2\%) & 75,963 (42.8\%) & 70,094 (41.5\%) \\
3 & 841,013 (37.6\%) & 71,799 (40.5\%) & 70,571 (41.8\%) \\
4 & 173,818 (7.8\%) & 14,247 (8.0\%) & 14,329 (8.5\%) \\
5 & 67,792 (3.0\%) & 5,446 (3.1\%) & 5,462 (3.2\%) \\
\midrule
\multicolumn{4}{l}{\textbf{Specialty}} \\
Cardiac & 34,918 (1.6\%) & 2,588 (1.5\%) & 2,673 (1.6\%) \\
ENT & 110,822 (4.9\%) & 8,495 (4.8\%) & 7,867 (4.7\%) \\
General Surgery & 473,207 (21.1\%) & 39,605 (22.3\%) & 41,071 (24.3\%) \\
Neurosurgery & 28,062 (1.3\%) & 2,416 (1.4\%) & 2,363 (1.4\%) \\
Obstetrics \& Gynaecology & 523,230 (23.4\%) & 40,621 (22.9\%) & 42,851 (25.4\%) \\
Ophthalmology & 29,902 (1.3\%) & 2,309 (1.3\%) & 2,320 (1.4\%) \\
Oral \& Maxillofacial & 14,896 (0.7\%) & 1,027 (0.6\%) & 1,065 (0.6\%) \\
Orthopaedics & 598,998 (26.7\%) & 46,002 (25.9\%) & 34,776 (20.6\%) \\
Other & 461 (0.0\%) & 54 (0.0\%) & 47 (0.0\%) \\
Plastics & 243,167 (10.9\%) & 18,470 (10.4\%) & 17,530 (10.4\%) \\
Thoracic & 13,015 (0.6\%) & 1,213 (0.7\%) & 1,207 (0.7\%) \\
Transplant & 3,617 (0.2\%) & 321 (0.2\%) & 346 (0.2\%) \\
Urology & 128,868 (5.8\%) & 11,637 (6.6\%) & 11,750 (7.0\%) \\
Vascular & 36,278 (1.6\%) & 2,637 (1.5\%) & 2,884 (1.7\%) \\
\midrule
\multicolumn{4}{l}{\textbf{Sub-specialty}} \\
Amputations & 7,559 (0.3\%) & 709 (0.4\%) & 629 (0.4\%) \\
Aortic & 6,963 (0.3\%) & 511 (0.3\%) & 569 (0.3\%) \\
Aortic Root / Ascending Aorta & 2,477 (0.1\%) & 227 (0.1\%) & 273 (0.2\%) \\
Arthroscopy & 124,054 (5.5\%) & 7,611 (4.3\%) & 1,954 (1.2\%) \\
Bladder & 28,485 (1.3\%) & 2,555 (1.4\%) & 2,470 (1.5\%) \\
Breast & 49,659 (2.2\%) & 4,145 (2.3\%) & 4,194 (2.5\%) \\
Burns & 3,312 (0.1\%) & 271 (0.2\%) & 253 (0.1\%) \\
Carotid & 4,039 (0.2\%) & 273 (0.2\%) & 291 (0.2\%) \\
Cervical \& Vulval & 26,872 (1.2\%) & 1,740 (1.0\%) & 2,093 (1.2\%) \\
Colorectal & 177,978 (7.9\%) & 14,896 (8.4\%) & 15,300 (9.1\%) \\
Coronary Artery Bypass (CABG) & 18,117 (0.8\%) & 1,287 (0.7\%) & 1,300 (0.8\%) \\
Cranial & 25,709 (1.1\%) & 2,207 (1.2\%) & 2,157 (1.3\%) \\
Ear Surgery & 18,559 (0.8\%) & 1,516 (0.9\%) & 1,481 (0.9\%) \\
Endocrine & 15,936 (0.7\%) & 1,316 (0.7\%) & 1,368 (0.8\%) \\
Flaps \& Reconstructions & 13,620 (0.6\%) & 1,007 (0.6\%) & 1,055 (0.6\%) \\
Fracture / Trauma Fixation & 144,099 (6.4\%) & 12,896 (7.3\%) & 12,361 (7.3\%) \\
HPB & 78,565 (3.5\%) & 7,693 (4.3\%) & 8,220 (4.9\%) \\
Hand Surgery & 66,808 (3.0\%) & 4,871 (2.7\%) & 4,163 (2.5\%) \\
Head \& Neck Tumour / Resection & 7,881 (0.4\%) & 653 (0.4\%) & 698 (0.4\%) \\
Heart & 156 (0.0\%) & 14 (0.0\%) & 14 (0.0\%) \\
Hernia/Abdominal Wall & 80,977 (3.6\%) & 6,421 (3.6\%) & 6,572 (3.9\%) \\
Hysterectomy \& Uterus & 152,780 (6.8\%) & 10,959 (6.2\%) & 11,323 (6.7\%) \\
Joint Replacement & 151,152 (6.7\%) & 12,922 (7.3\%) & 12,468 (7.4\%) \\
Kidney & 11,988 (0.5\%) & 1,346 (0.8\%) & 1,402 (0.8\%) \\
Liver & 450 (0.0\%) & 43 (0.0\%) & 44 (0.0\%) \\
Lung & 237 (0.0\%) & 22 (0.0\%) & 28 (0.0\%) \\
Lung Resection & 5,722 (0.3\%) & 634 (0.4\%) & 597 (0.4\%) \\
Mediastinal & 624 (0.0\%) & 41 (0.0\%) & 46 (0.0\%) \\
Mesenteric / Visceral & 215 (0.0\%) & 14 (0.0\%) & 20 (0.0\%) \\
Miscellaneous Abdominal / Peritoneal & 54,356 (2.4\%) & 3,839 (2.2\%) & 3,945 (2.3\%) \\
Nasal / Sinus Surgery & 42,161 (1.9\%) & 3,168 (1.8\%) & 2,451 (1.5\%) \\
Obstetrics & 281,411 (12.6\%) & 23,962 (13.5\%) & 25,703 (15.2\%) \\
Ophthalmology & 29,902 (1.3\%) & 2,309 (1.3\%) & 2,320 (1.4\%) \\
Oral \& Maxillofacial & 14,896 (0.7\%) & 1,027 (0.6\%) & 1,065 (0.6\%) \\
Other & 461 (0.0\%) & 54 (0.0\%) & 47 (0.0\%) \\
Other Cardiac & 1,382 (0.1\%) & 134 (0.1\%) & 129 (0.1\%) \\
Other ENT & 13,013 (0.6\%) & 1,051 (0.6\%) & 1,013 (0.6\%) \\
Other Gynaecology & 12,836 (0.6\%) & 1,043 (0.6\%) & 973 (0.6\%) \\
Other Neurosurgery & 150 (0.0\%) & 13 (0.0\%) & 15 (0.0\%) \\
Other Orthopaedics & 120,064 (5.4\%) & 8,065 (4.5\%) & 4,501 (2.7\%) \\
Other Soft Tissue & 148,753 (6.6\%) & 11,569 (6.5\%) & 11,345 (6.7\%) \\
Other Thoracic & 3,037 (0.1\%) & 266 (0.1\%) & 286 (0.2\%) \\
Other Urology & 1,309 (0.1\%) & 77 (0.0\%) & 36 (0.0\%) \\
Other Vascular & 3,171 (0.1\%) & 565 (0.3\%) & 686 (0.4\%) \\
Ovarian \& Adnexal & 31,238 (1.4\%) & 2,162 (1.2\%) & 2,106 (1.2\%) \\
Pelvic Floor \& Prolapse & 18,093 (0.8\%) & 755 (0.4\%) & 653 (0.4\%) \\
Peripheral Arterial & 15,383 (0.7\%) & 1,046 (0.6\%) & 1,070 (0.6\%) \\
Peripheral Nerve & 238 (0.0\%) & 16 (0.0\%) & 11 (0.0\%) \\
Pleural Procedures & 3,632 (0.2\%) & 272 (0.2\%) & 278 (0.2\%) \\
Prostate & 31,896 (1.4\%) & 2,694 (1.5\%) & 2,817 (1.7\%) \\
Scar Revision \& Cosmetic & 10,674 (0.5\%) & 752 (0.4\%) & 714 (0.4\%) \\
Spine & 54,035 (2.4\%) & 3,979 (2.2\%) & 3,043 (1.8\%) \\
Testis / Scrotum / Penis & 23,487 (1.0\%) & 1,756 (1.0\%) & 1,656 (1.0\%) \\
Throat / Larynx / Airway & 29,208 (1.3\%) & 2,107 (1.2\%) & 2,224 (1.3\%) \\
Upper GI & 15,739 (0.7\%) & 1,295 (0.7\%) & 1,473 (0.9\%) \\
Ureter & 34,474 (1.5\%) & 3,451 (1.9\%) & 3,628 (2.1\%) \\
Valve Surgery & 12,942 (0.6\%) & 940 (0.5\%) & 971 (0.6\%) \\
Venous & 6,507 (0.3\%) & 228 (0.1\%) & 248 (0.1\%) \\
\bottomrule
\end{longtable}

\end{document}